\documentclass[10pt,twocolumn,letterpaper]{article}

\usepackage{iccv}
\usepackage{times}
\usepackage{epsfig}
\usepackage{graphicx}
\usepackage{amsmath}
\usepackage{amssymb}
\usepackage{booktabs}
\newtheorem{definition}{Definition}
\usepackage{booktabs}
\usepackage{float}
\usepackage[table]{xcolor}
\usepackage{lipsum}

\usepackage[pagebackref=true,breaklinks=true,letterpaper=true,colorlinks,bookmarks=false]{hyperref}
\iccvfinalcopy 

\def\iccvPaperID{9683} 

\ificcvfinal\fi

\begin{document}

\title{Shape-Biased Domain Generalization via Shock Graph Embeddings}

\author{Maruthi Narayanan, Vickram Rajendran\\
Johns Hopkins University Applied Physics Laboratory\\
Laurel, MD 20723\\
{\tt\small \{Maruthi.Narayanan, Vickram.Rajendran\}@jhuapl.edu}
\and
Benjamin Kimia\\
Brown University\\
School of Engineering\\
{\tt\small benjamin\_kimia@brown.edu}
}

\maketitle
\ificcvfinal\thispagestyle{empty}\fi

\begin{abstract}

  There is an emerging sense that the vulnerability of Image Convolutional Neural Networks (CNN), i.e., sensitivity to image corruptions, perturbations, and adversarial attacks, is connected with Texture Bias. This relative lack of Shape Bias is also responsible for poor performance in Domain Generalization (DG). The inclusion of a role of shape alleviates these vulnerabilities and some approaches have achieved this by training on negative images, images endowed with edge maps, or images with conflicting shape and texture information. This paper advocates an explicit and complete representation of shape using a classical computer vision approach, namely, representing the shape content of an image with the shock graph of its contour map. The resulting graph and its descriptor is a complete representation of contour content and is classified using recent Graph Neural Network (GNN) methods. The experimental results on three domain shift datasets, Colored MNIST, PACS, and VLCS demonstrate that even without using appearance the shape-based approach exceeds classical Image CNN based methods in domain generalization.
   
\end{abstract}

\section{Introduction}

The incredible success of deep learning methods has outpaced our understanding of how they work. An avalanche of new Convolutional Neural Networks (CNN) methods, many with computer vision applications, have been developed in the context of improving performance without a simultaneous development of fundamental understanding of its mechanisms. This has side-lined an integration with the traditional classical AI/computer vision approaches and cognitive science studies. The insight from the classical methods on representation and search is ignored mainly because it is not clear how to integrate the two paradigms: one of developing representations based on feedback from degree of success~\cite{LeCun:etal:NATURE15}, and one on developing representations based on the more classic geometric and world scene reasoning~\cite{Marr:Visual:1978}.

Hand in hand with the continued success of CNNs in difficult problems such as image classification~\cite{Munandet:etal:ICML13,Ghifary:etal:CVPR15,Li:etal:CVPR18,Li:etal:ECCV18,Motiian:etal:ICCV17,Li:etal:ICCV17,Khosla:etal:ECCV12,Carlucci:etal:CVPR19,Shankar:etal:ICLR18,Volpi:etal:NIPS18,Balaji:etal:NIPS2018,Li:etal:AAAI18,Li:etal:ICCV19,Dou:etal:NIPS19,Qiao:etal:CVPR20}, their vulnerabilities have also been identified. Image CNNs struggle with ``domain shift'' \ie, when the training conditions (resolutions, sensors, \etc) do not match the conditions of the operating domain. Small corruptions \eg, noise, blur, fog, \etc, and perturbations which do not affect human perception~\cite{Azulay:Weiss:JMLR19} affect the performance of Image CNNs as benchmarked by~\cite{Hendrycks:etal:ICLR19}. In particular, when these small changes are adversarial, \ie, targeted to the architecture of the CNN for maximal effect~\cite{szegedy2013intriguing,goodfellow2014explaining,moosavi2016deepfool,carlini2017towards}, the performance can be seriously affected (adversarial attack).


There is an emerging understanding that underlying these vulnerabilities is a lack of representation of shape. While initially it was argued, \eg, by visualizations in~\cite{Zeiler:Fergus:ECCV14}, that Image CNNs construct low-level primitives such as corners and edges in lower layers, which in turn lead to more complex shapes in higher layers~\cite{Ritter:etal:ICML17,LeCun:etal:NATURE15,Kreiegeskorte:etal:ARVS15} there is increasing evidence that appearance (texture bias) and not global shape (shape bias) is the cornerstone of correct classification by CNNs~\cite{Brendel:Bethge:ICLR19,Shi:etal:ICML20}. Baker et al~\cite{Baker:etal:PLOS18} explored the role of object shape in CNNs trained for image classification and found that the global object shape, which is arguably the most important cue for human categorization of objects, is not represented. Similarly, Geirhos \etal~\cite{Geirhos:etal:ICLR19} show that Image CNNs only rely on object texture rather than object shape and show how with proper training the introduction of a shape bias improves performance. In the field of cognitive science, Malhotra and Bowers~\cite{Malhotra:Bowers:CS19} systematically explore the impact of non-shape features in the categorization performance of CNNs on the CIFAR-10~\cite{Krizhevsky09learningmultiple} dataset. They conclude that even though the CNN mimics the hierarchical architecture and learning processes of biological vision, they pick up optimal features to perform prediction which do not include shape. They conjecture that the lack of an intrinsic shape bias in a typical Image CNN architecture can be the key to susceptibility to adversarial attacks. Zhang and Zhu~\cite{Zhang:etal:CVPR20} find that adversarially trained networks are less texture-biased and more shape-biased. More recently~\cite{Baker:etal:VR20} shows that CNNs performance is an inversion of human performance with respect to global and local shape priors in that Image CNNs capture only local shape information while humans capture absolute shape information. See also~\cite{Carlucci:etal:CVPR19,Wang:etal:NIPS19}. 

\begin{figure}[ht]
\center
{\footnotesize\textit{\textcolor{black}{a)}}}\includegraphics[width=0.29\textwidth]{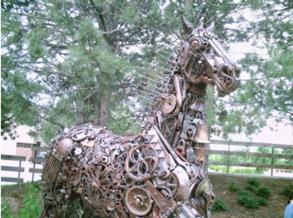}
{\footnotesize\textit{\textcolor{black}{b)}}}\includegraphics[width=0.12\textheight]{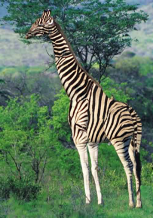}
\caption{Observe that shape is dominant in forming perception when object features have not been experienced before (a), or are completely contradictory where the percept is a giraffe and not a zebra (b)~\cite{Trinh:Kimia:IJCV11}. }
\label{fig:horse_examples}
\end{figure}

\begin{figure}[ht]
\center
{\footnotesize\textit{\textcolor{black}{a)}}}\includegraphics[width=0.112\textwidth]{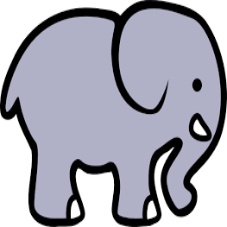}
\includegraphics[width=0.112\textwidth]{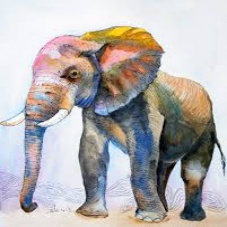}
\includegraphics[width=0.112\textwidth]{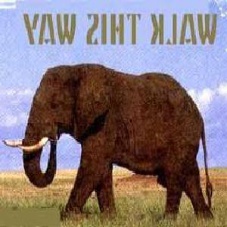}
\includegraphics[width=0.112\textwidth]{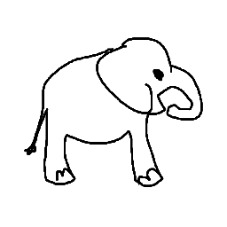}
{\footnotesize\textit{\textcolor{black}{b)}}}\includegraphics[width=0.112\textwidth]{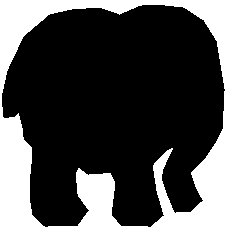}
\includegraphics[width=0.112\textwidth]{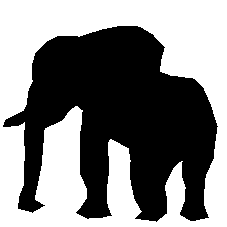}
\includegraphics[width=0.112\textwidth]{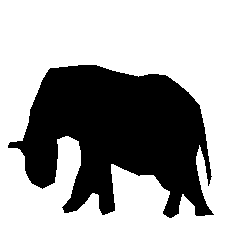}
\includegraphics[width=0.112\textwidth]{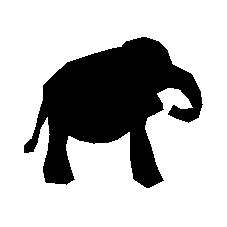}
{\footnotesize\textit{\textcolor{black}{c)}}}\includegraphics[width=0.112\textwidth]{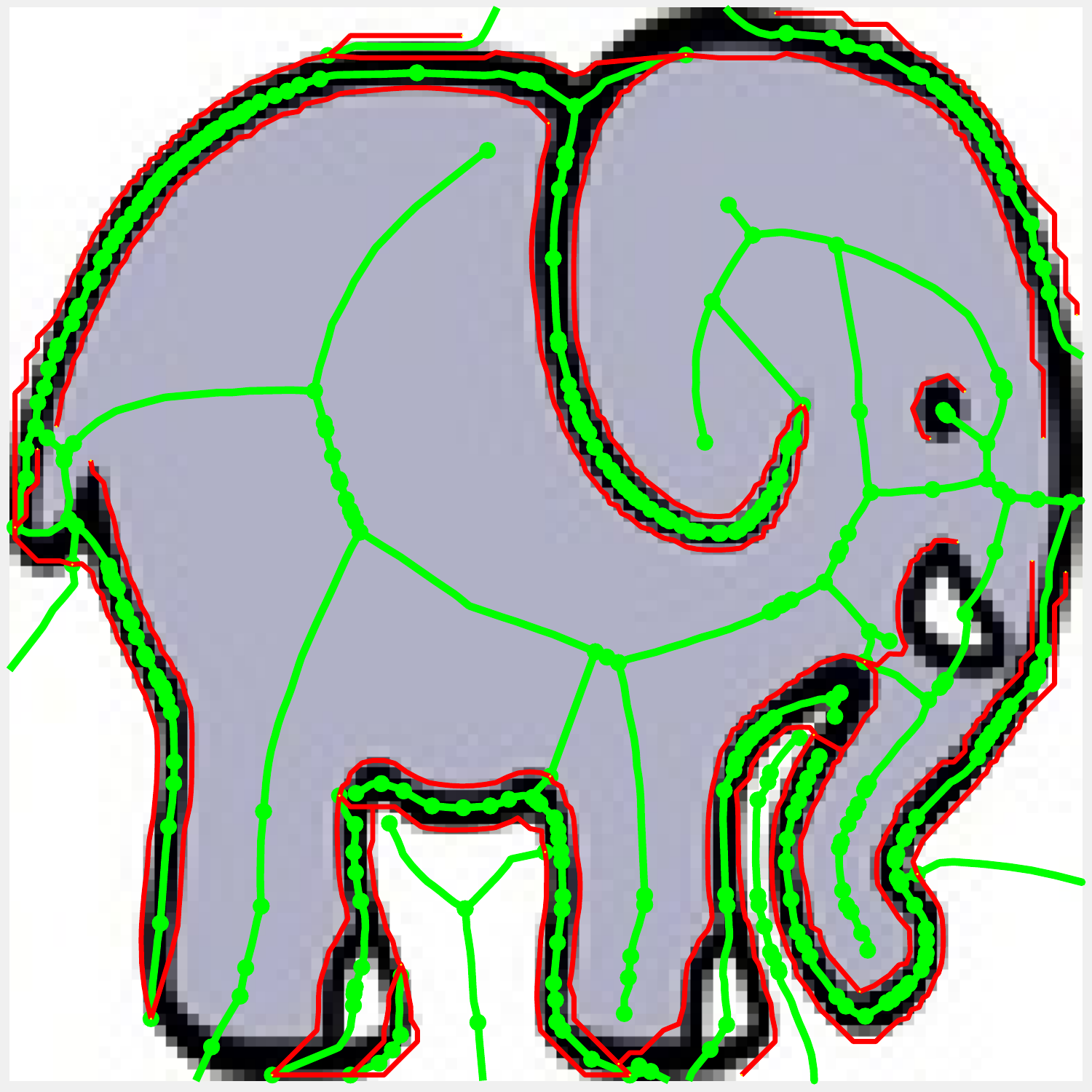}
\includegraphics[width=0.112\textwidth]{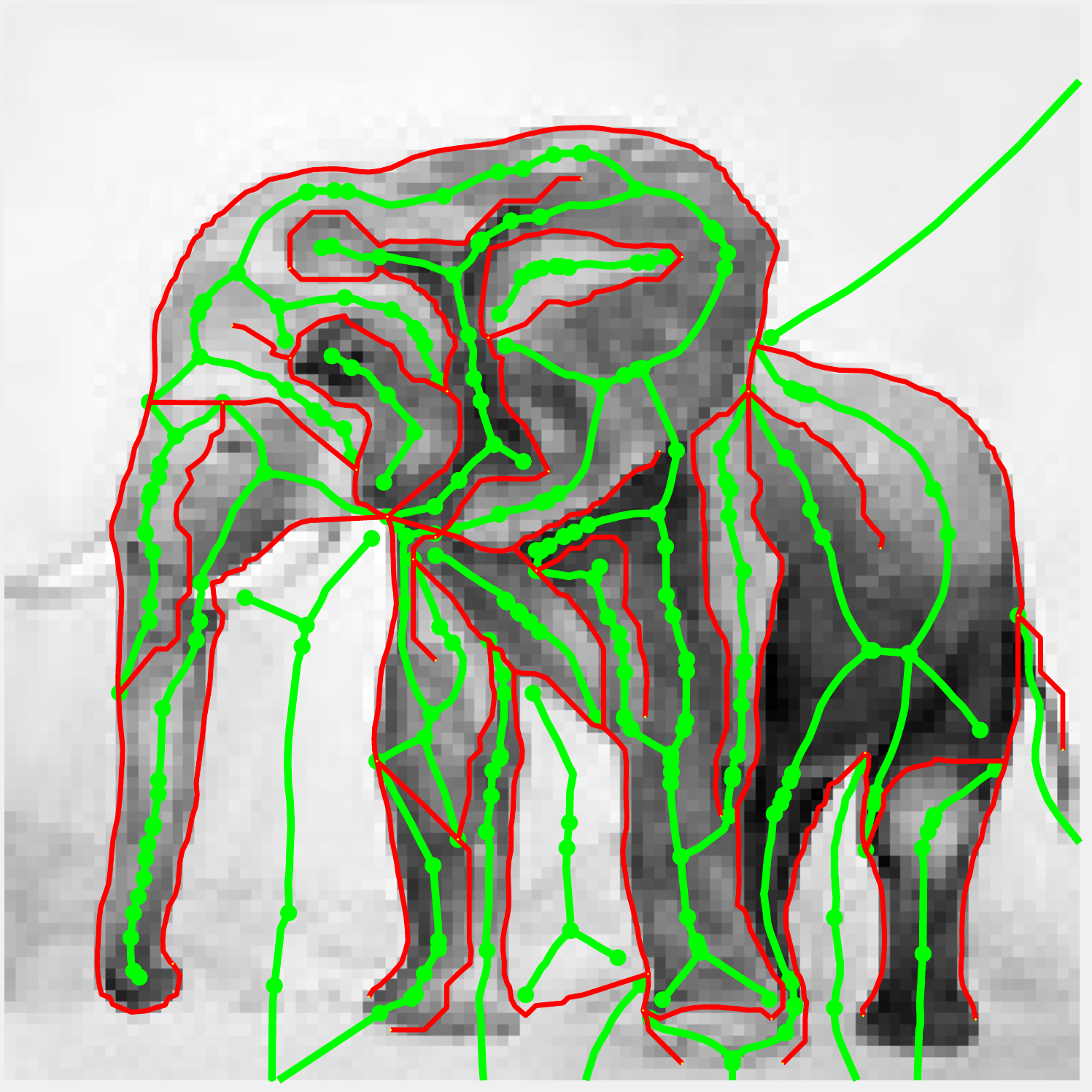}
\includegraphics[width=0.112\textwidth]{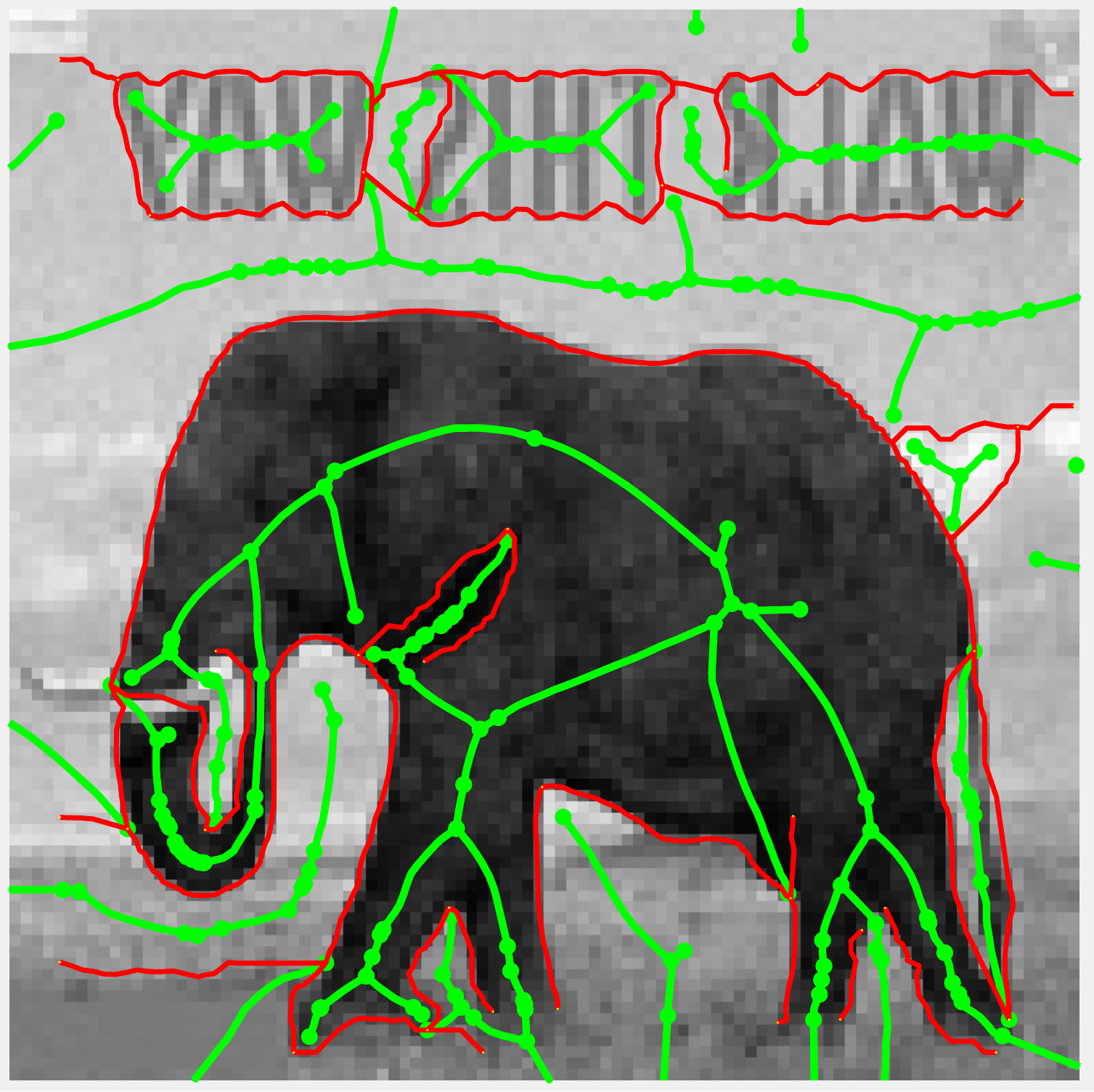}
\includegraphics[width=0.112\textwidth]{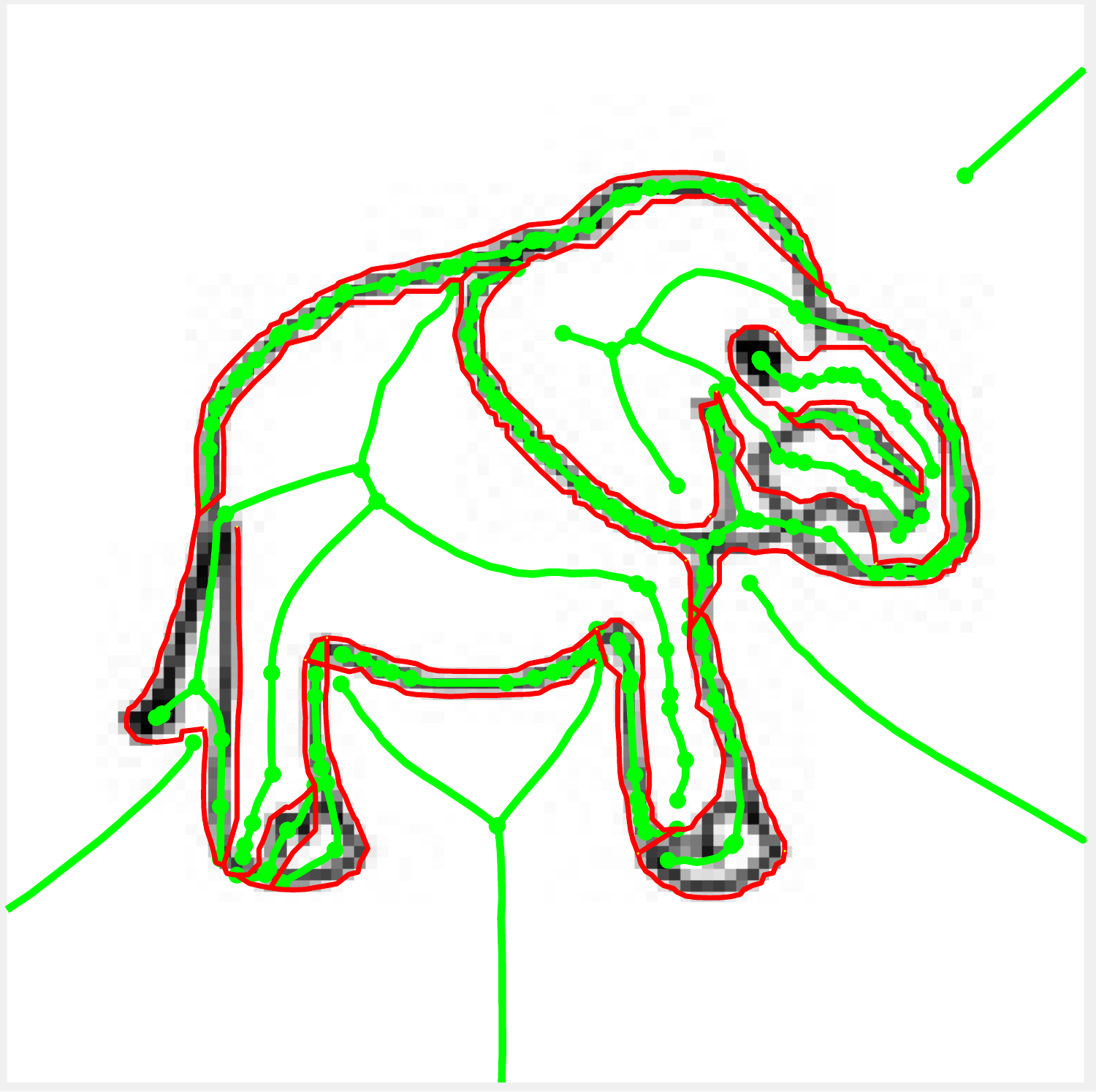}

\caption{a) Various images of an elephant across the four domains from the PACS~\cite{Li:etal:ICCV17} dataset. b) The shape of an object as captured by a binary mask of the images. c) The shock graph of the contour map colored in \textcolor{green}{green} (shock dynamics are not shown). We encourage readers to zoom in.}

\label{fig:sg_motivate}
\end{figure}

Shape is an object constancy that dominates object texture, Figure~\ref{fig:horse_examples}, and is invariant to image corruption or perturbation. Observe in Figure~\ref{fig:sg_motivate}\textcolor{red}{a}, that the objects in the images taken from standard domain generalization datasets exhibit significant variation in color and/or texture. Observe that after delineation of the object of interest, Figure~\ref{fig:sg_motivate}\textcolor{red}{b}, the variation in shape is much less than the appearance. This concept of ``shape constancy'' has long been a pillar of classic computer vision and recognized as one of the most important cues in biological object recognition~\cite{Biederman:RBC,Biederman:Ju:1988}. We conjecture that the strong bias of Image CNNs to texture cues coupled with the inability to capture global shape cues is limiting their capability to generalize. Several researchers have aimed to increase the shape bias of CNNs, for example by training on negative images which preserve shape but not appearance~\cite{Hosseini:etal:CVPRW18}. Borji augments the training by including edge maps~\cite{Borji:ARXIV20}. Li \etal~\cite{Li:etal:ICLR21} augment training data with images with conflicting shape and texture information, \eg, an image of chimpanzee shape but with lemon texture, under simultaneous shape and texture supervision to develop each independently. It is clear from these studies that the enhancing and emphasizing the shape cue in the training process improves performance.

This paper advocates a complete and explicit representation of shape. Instead of forcing the Image CNN to learn shape indirectly, shape is directly extracted and represented based on the most advanced understanding of classical computer vision. In other words, each image is augmented with an explicit representation of its shape content. In fact, the image itself is discarded for the purpose of this study and we solely train on the explicit shape representation to completely characterize how far image classification can proceed on a shape-only platform. 

The segmented images in Figure~\ref{fig:sg_motivate}\textcolor{red}{b} represent the canonical definition of ``shape'' in the computer vision community. Shape descriptors aim to capture regional and/or boundary properties of this binary shape. A sampling of popular descriptors include shape context~\cite{Mori:etal:PAMI05}, medial axis descriptors~\cite{Giblin:Kimia:MedialBook07,Vidal:etal:MedialAxis:ICPR00}, and shape tree~\cite{Felzenszwalb:Schwartz:CVPR:2007}, just to name a few among many. The vast majority of approaches then rely on some form of assignment or graph matching algorithms to define a distance, and that distance is subsequently used in shape classification pipelines. The medial axis~\cite{Blum67Transformation} of a segmented shape is a particular favorite as it captures invariances for many deformations.

Since segmentation itself is a difficult problem, the notion of shape must be surmised directly from edges or regions. The shock graph~\cite{Kimia:etal:ECCV:Book,Kimia:etal:Shape:Series:I,Giblin:Kimia:Reconstruction:PAMI03}, a refinement of medial axis descriptors, has made the leap from binary images to real images~\cite{Narayanan:Kimia:ECCV12,Ozcanli:Kimia:BMVC07}. Observe in Figure~\ref{fig:sg_motivate}\textcolor{red}{c}, the shock graph of the original images, Figure~\ref{fig:sg_motivate}\textcolor{red}{a}, captures the ``skeleton'' of these various elephants. We can see the delineation of the various common parts of the elephant, distinctive trunk and legs, shared among all elephant exemplars despite originating from disparate domains. Further motivating factors for why we use the shock graph is that it is robust to articulations and occlusion of the underlying shape which are more common in realistic image acquisition settings. Its state of the art performance on standard shape datasets~\cite{Sebastian:etal:Shocks:PAMI2004} is an additional reason to use it in this research.

The shock graph is a complete representation of image contours in that they can be reconstructed back from the shock graph, which encodes both the axis and the dynamics along the axis. In this sense the shock graph is a complete representation of the object contours embedded in an image. Thus, training can take place on the shock graph instead of the original image. Observe, however, that the image pixel topology is now replaced with a graph topology. Recent advancements in Graph Neural Networks (GNN)~\cite{Wu:etal:Yu:Survey:GNN,Zhou:etal:Sun:Review:GNN} can capture the information in the graph for the purpose of classification. Specifically, our pipeline is one of transforming the image to an edge map, a contour map, and then a shock graph which is subsequently trained on using a graph neural network for classification. This represents a novel use of graph neural networks as the vast majority of applications of graph neural networks have focused on graph machine learning tasks (node labeling, edge labeling, and graph classification) predominantly for social networks or in chemistry with molecule graphs. We show that this representation of shape information via a shock graph and using GNN to train on them leads to excellent classification performance, even though the appearance is not taken into account. Future work will likely enhance this performance further when both cues are utilized.

Our contributions can be summarized as follows. \emph{1)} To the best of our knowledge the first paper to revisit classic shape descriptors in a modern setting for domain generalization \emph{2)} A novel approach of using an image-to-graph transformation instead of the traditional Image CNN input. \emph{3)} Combining shock-graphs with graph neural networks to achieve state of the art graph classification performance comparable to the image classification performance achieved by standard Image CNN architectures. \emph{4)} A novel image-to-graph framework that shows the power of the shape bias, with the potential to integrate texture cues. From a high-level perspective, the paper also shows that it is possible to bring to bear lessons learned from classical computer vision in the context of modern deep learning techniques.

\section{Related Work}

Domain generalization approaches aim to generalize models to unseen target domains without having any prior information about the target distribution during training.  Due to the challenging nature of this problem a wide variety of methods have been tried. From a high-level, DG methods can roughly be organized into one of three families: \emph{1)} domain invariant features \emph{2)} data augmentation and finally \emph{3)} optimization algorithms. We do not claim this taxonomy exhaustive or unique, as many methods may fall into one or more or none of the paradigms, but rather as a way to cover the vast majority of research in this field. 

Learning domain invariant features~\cite{Munandet:etal:ICML13,Ghifary:etal:CVPR15,Li:etal:CVPR18,Li:etal:ECCV18,Motiian:etal:ICCV17,Li:etal:ICCV17,Khosla:etal:ECCV12} represents the most popular approach to the problem. The intuition being that if we can learn a domain-invariant representation across multiple disparate source domains that this representation will translate well to the target too. Methods such as~\cite{Li:etal:CVPR18,Munandet:etal:ICML13} develop learning algorithms that optimize the maximum mean discrepancy (MMD) or variations of it to minimize the feature dissimilarity across multiple domains. Similar to those approaches~\cite{Li:etal:ECCV18} also tries to regularize the features space, but utilizes a probabilistic framework to minimize the joint discrepancy of labels and inputs across multiple domains. The contrastive loss is used in~\cite{Motiian:etal:ICCV17} to encourage features from similar classes across multiple domains to cluster in the latent space. The work of~\cite{Ghifary:etal:CVPR15} on the other hand utilizes a multi-task auto encoder to directly extract out invariant features robust enough to domain changes. The  previous approaches relied on some form of feature-space regularization across domains, while the works of~\cite{Li:etal:ICCV17,Khosla:etal:ECCV12} alter the model architecture to achieve invariance. \cite{Li:etal:ICCV17} uses a low-rank parametrized CNN. The approach of~\cite{Khosla:etal:ECCV12} decomposes ML models into their domain specific and invariant components to reduce domain shift.

Data augmentation techniques~\cite{Carlucci:etal:CVPR19,Shankar:etal:ICLR18,Volpi:etal:NIPS18,Qiao:etal:CVPR20} rely on the underlying assumption that more data will lead to more generalizable models and less overfitting. ~\cite{Shankar:etal:ICLR18} generates new data samples through gradient perturbations, while~\cite{Volpi:etal:NIPS18} generates adversarial perturbed samples. The recent work of~\cite{Carlucci:etal:CVPR19} effectively increases the data by viewing the domain generalization problem of one as puzzle solving where the pieces are the various numerous squares within the source domain images.   

Optimization approaches~\cite{Balaji:etal:NIPS2018,Li:etal:AAAI18,Li:etal:ICCV19,Dou:etal:NIPS19} aim to modify traditional learning schemes in domain generalization to achieve a more robust solution during training. The most popular approach is to leverage meta-learning for domain generalization. Specifically the episodic training paradigm, originating from model-agnostic meta-learning (MAML)~\cite{Fin:etal:ICML17} has been repurposed to specifically address domain generalization in~\cite{Balaji:etal:NIPS2018,Li:etal:AAAI18,Li:etal:ICCV19,Dou:etal:NIPS19}.  The work ~\cite{Li:etal:AAAI18} can be seen as direct adaptation of MAML~\cite{Fin:etal:ICML17} for domain generalization. While the approach of~\cite{Li:etal:ICCV19} uses an episodic training scheme to alternate learning domain specific features/classifiers across domains, it ignores the inner-gradient update. In~\cite{Balaji:etal:NIPS2018} the feature extractor and classifier are treated separately where the classifier is regularized according to a learned function the feature extractor remains the same. The approach of~\cite{Dou:etal:NIPS19} can be seen as hybrid approach as it incorporates semantic feature space knowledge into the meta-learning process.

There has been minimal work in the last few years as it relates to the shock graph.  Recent work has focused exclusively on computational algorithms to compute the shock graph/medial axis~\cite{Camaro:etal:CVPR20,Tsogkas:Dickinson:ICCV17} from real images with no further applications to computer vision tasks. Different from previous works, our is the first to combine the shock graph with recent advancements in graph based deep learning , with applications to domain generalization. Our approach to domain generalization falls under the invariant-feature approach. Given that our work proposes a new feature extractor it can be used in any family of methods \ie drop in replacement for Image CNN features. The other major distinction with our work compared to others is the use of Graph Neural Networks. We are not aware of any other works using GNNs in the DG area. We don’t claim a novel new GNN architecture but rather the novelty of applying to the problem of domain generalization.

\section{Shock Graph}
\label{sec:sg}

\begin{figure}[ht]
\center
\includegraphics[width=0.45\textwidth]{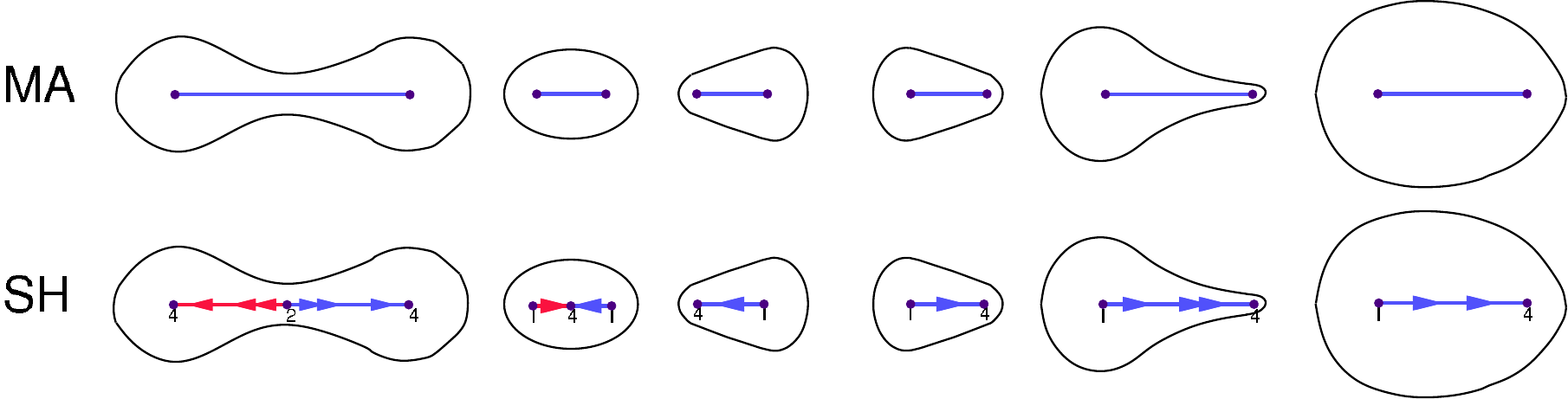}
\caption{A closed contour in black and the resulting medial axis and shock graph delineated in blue/red. From~\cite{Giblin:Kimia:IJCV03}. The advantage of the shock graph over medial axis is its qualitative description of the boundary: given a medial axis segment any of the six shapes in the top row can occur, while a shock graph segment (a monotonically flowing subsegment of the medial axis) qualitatively represents one type of shape.}
\label{fig:ma_vs_sg}
\end{figure}

The medial axis is a popular region-based description of shape. Figure~\ref{fig:ma_vs_sg} shows the medial axis of a simple closed contour, which is defined as the closure of the set of centers of the contour’s maximal circles. The shock graph~\cite{Kimia:etal:ECCV:Book,Kimia:etal:Shape:Series:I,Giblin:Kimia:Reconstruction:PAMI03}, a refinement of the Medial Axis, arises from viewing the medial axis as the locus of singularities (shocks) formed in the course of wave propagation (grass-fire) from contour boundaries, Figure~\ref{fig:ma_vs_sg}. During this wave propagation process, the edges and nodes that comprise the shock graph are formed. At each step, the interaction of waves emanating from contour elements quench at shock edges \emph{i.e.} bisector of curves. Shock edges are the valid portion of these infinite bisector curves, and nodes are formed where endpoints of curves interact and/or bisector curves intersect. Further details about the computational algorithm can be found in the supplementary material. The resulting shock graph, at the end of this process is a richer descriptor of the contour map than the medial axis as it allows us to refine the classification of nodes/edges according to the underlying dynamics of wave propagation as well as the presence or absence or shock edges/nodes which identify interacting contour pairs and triplets. This graph capturing the relationship between contours is our proposed “shape” cue for domain generalization. Formally,

\begin{definition}
\label{def:sg}
The shock graph is a directed attributed relational graph, $G=(V,E,A)$, where $V$ is the vertex set of shock nodes, $E$ is the edge set of shock segments, and $A$ is the continuous intrinsic geometry and classification labels, specifically $A$ is the set of unary attributes $a_i$ attached to each vertex $v_i \in V$, namely normal, tangent, time of formation and discrete labels sink, source, or junction. The shock link binary attributes $a_{ij}$ attached to each $e_k=(v_i,v_j) \in E$ consist of length, curvature, and acceleration and discrete labels degenerate, semi-degenerate, or regular. 
\end{definition}

\begin{figure}[!ht]
\centering
{\footnotesize\textit{\textcolor{black}{a)}}}\includegraphics[width=0.214\linewidth]{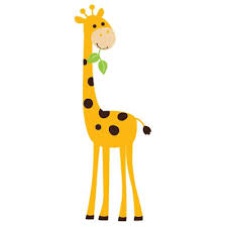} 
{\footnotesize\textit{\textcolor{white}{a)}}}\includegraphics[width=0.214\linewidth]{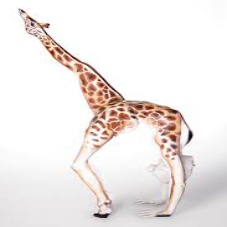} 
{\footnotesize\textit{\textcolor{white}{a)}}}\includegraphics[width=0.214\linewidth]{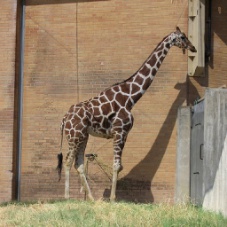} 
{\footnotesize\textit{\textcolor{white}{a)}}}\includegraphics[width=0.214\linewidth]{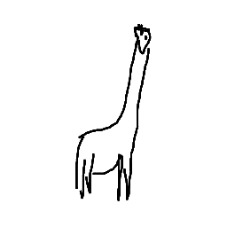}
{\footnotesize\textit{\textcolor{black}{b)}}}\includegraphics[width=0.214\linewidth]{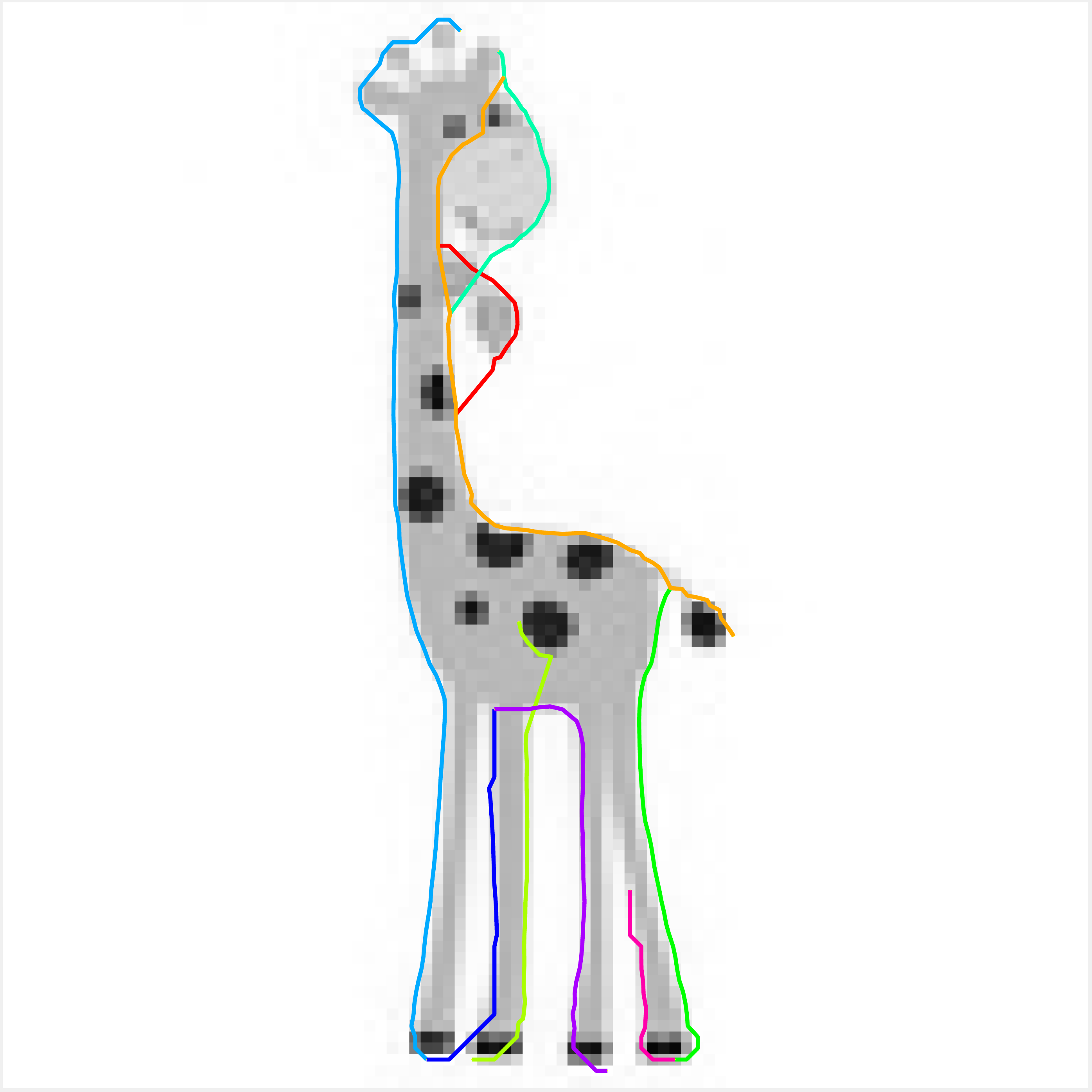}
{\footnotesize\textit{\textcolor{white}{a)}}}\includegraphics[width=0.214\linewidth]{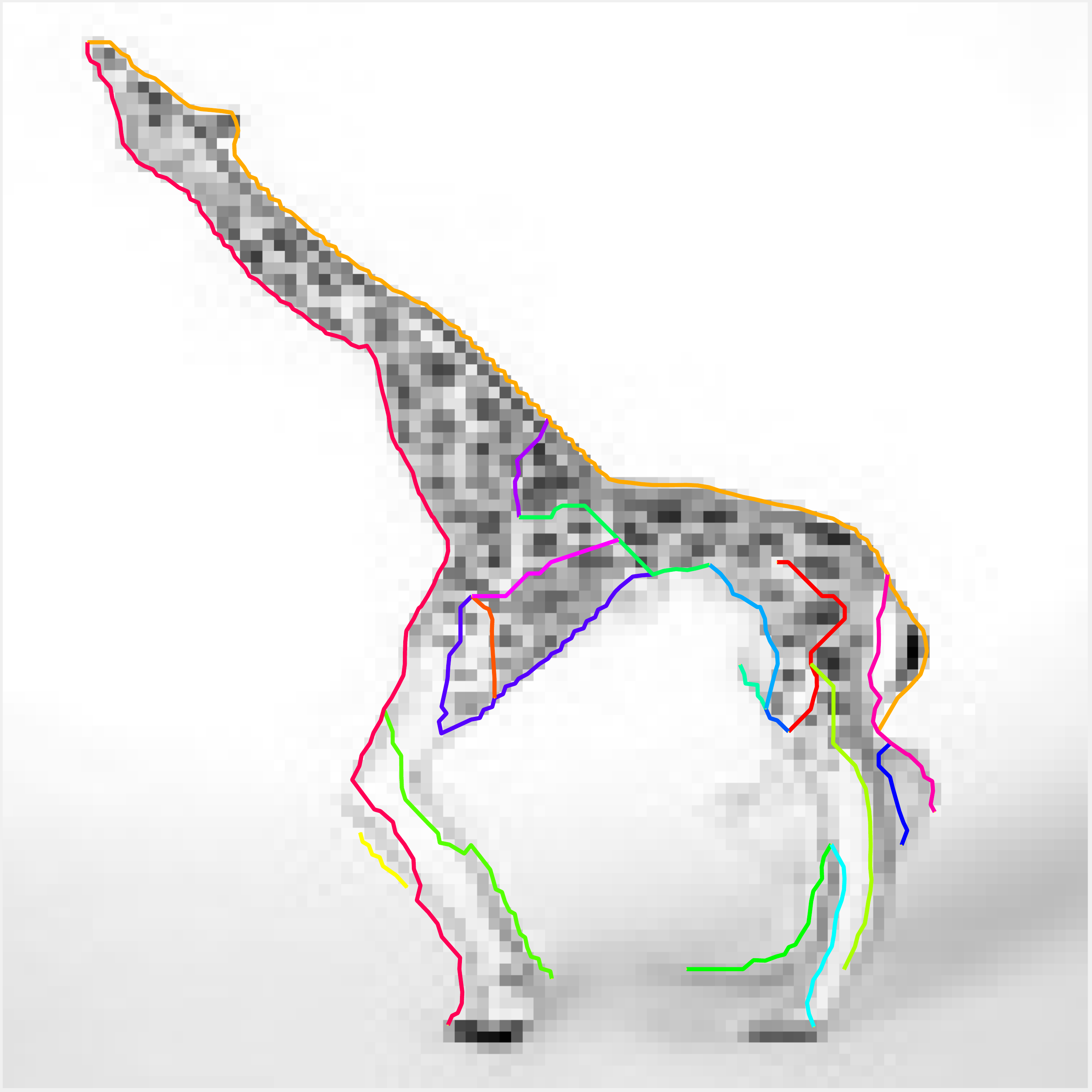}
{\footnotesize\textit{\textcolor{white}{a)}}}\includegraphics[width=0.214\linewidth]{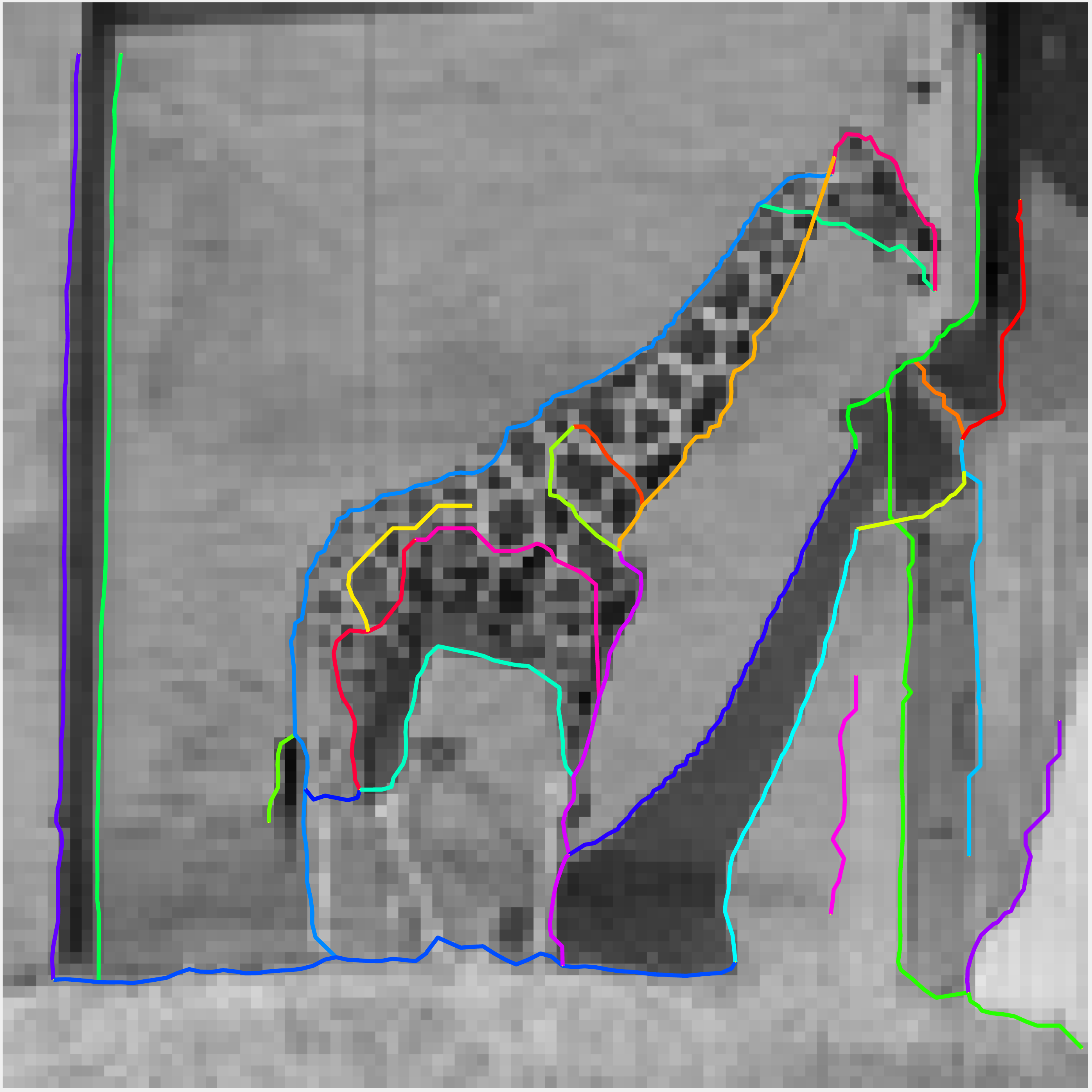}
{\footnotesize\textit{\textcolor{white}{a)}}}\includegraphics[width=0.214\linewidth]{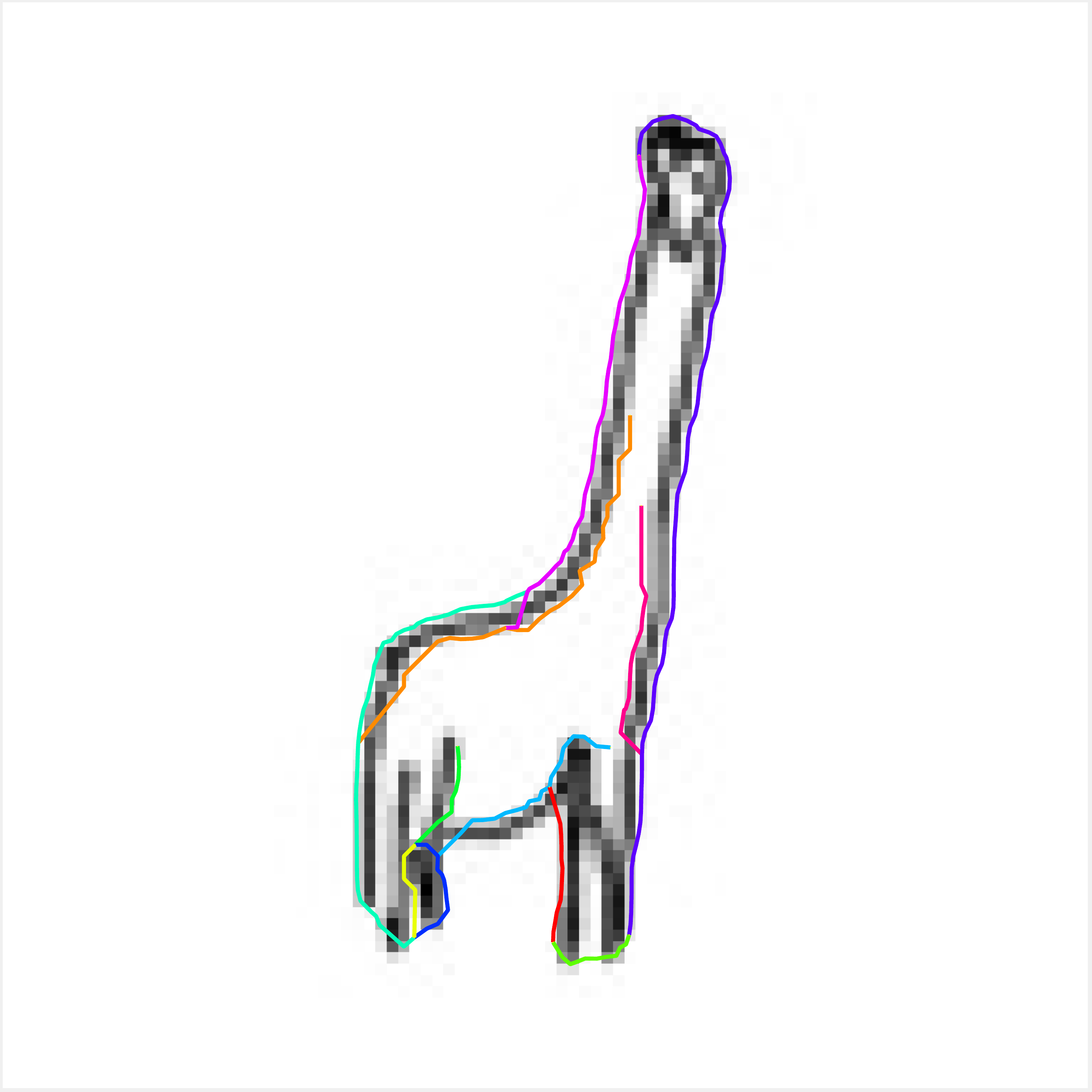}
{\footnotesize\textit{\textcolor{black}{c)}}}\includegraphics[width=0.214\linewidth]{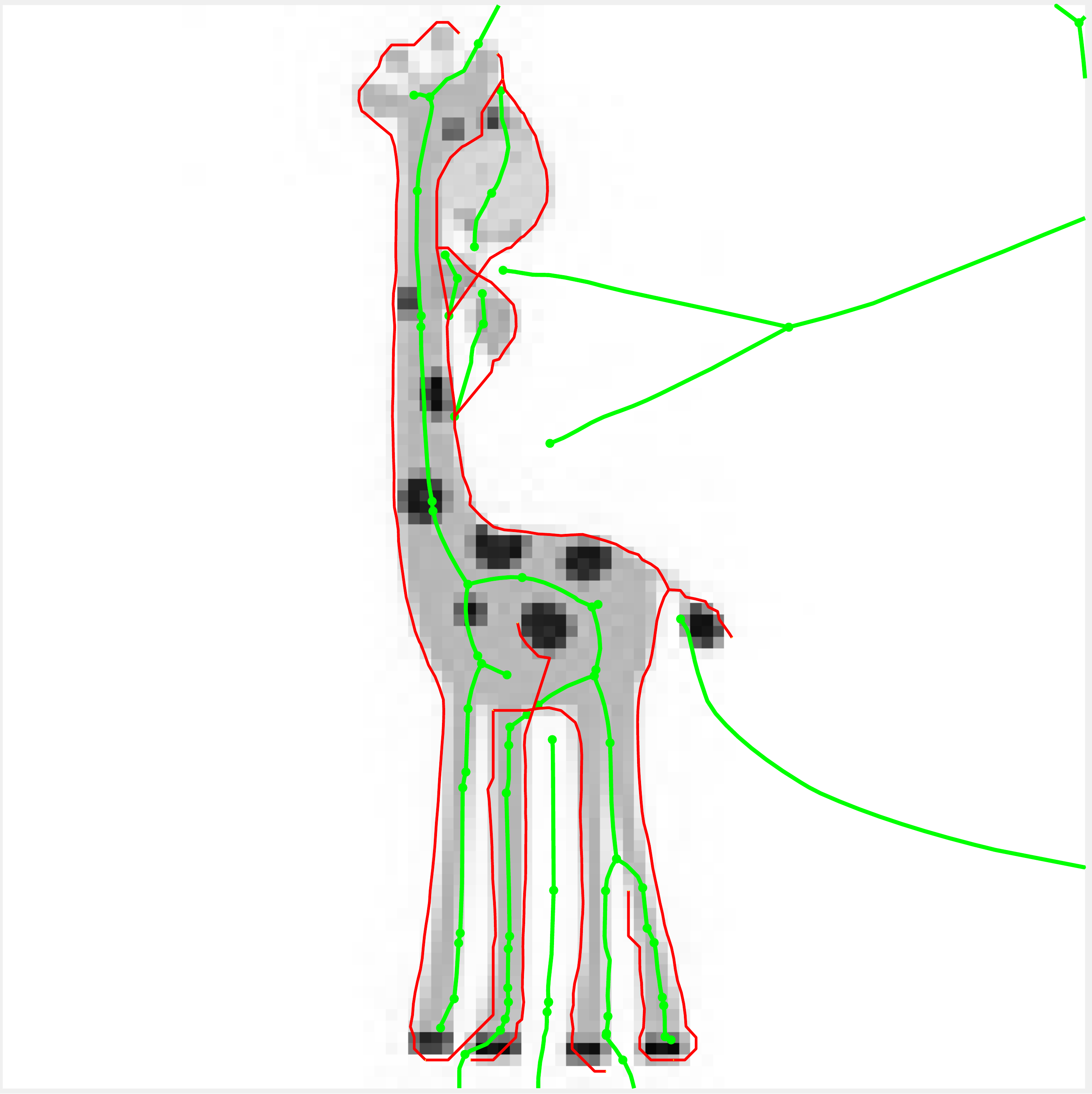}
{\footnotesize\textit{\textcolor{white}{a)}}}\includegraphics[width=0.214\linewidth]{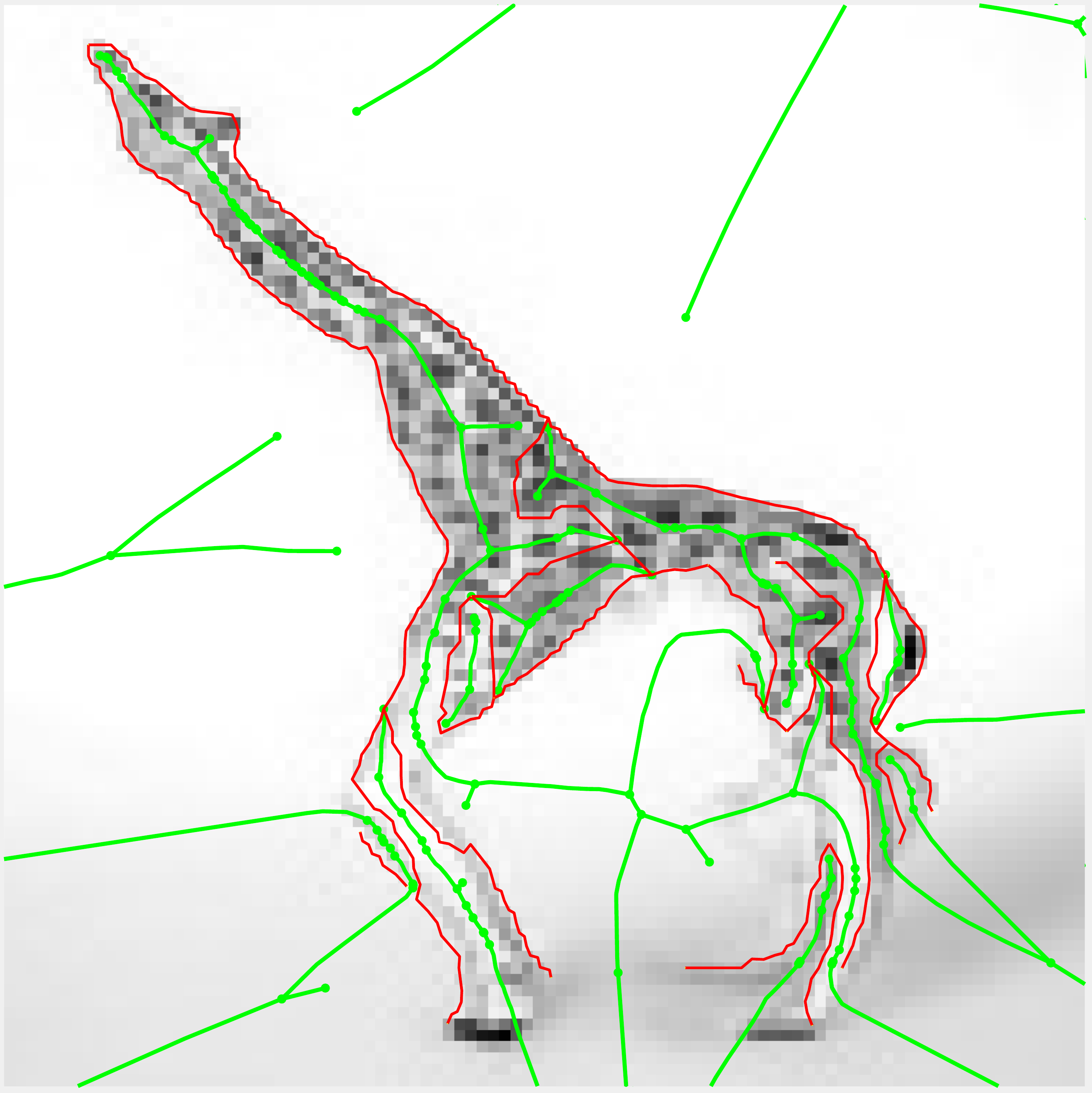} 
{\footnotesize\textit{\textcolor{white}{a)}}}\includegraphics[width=0.214\linewidth]{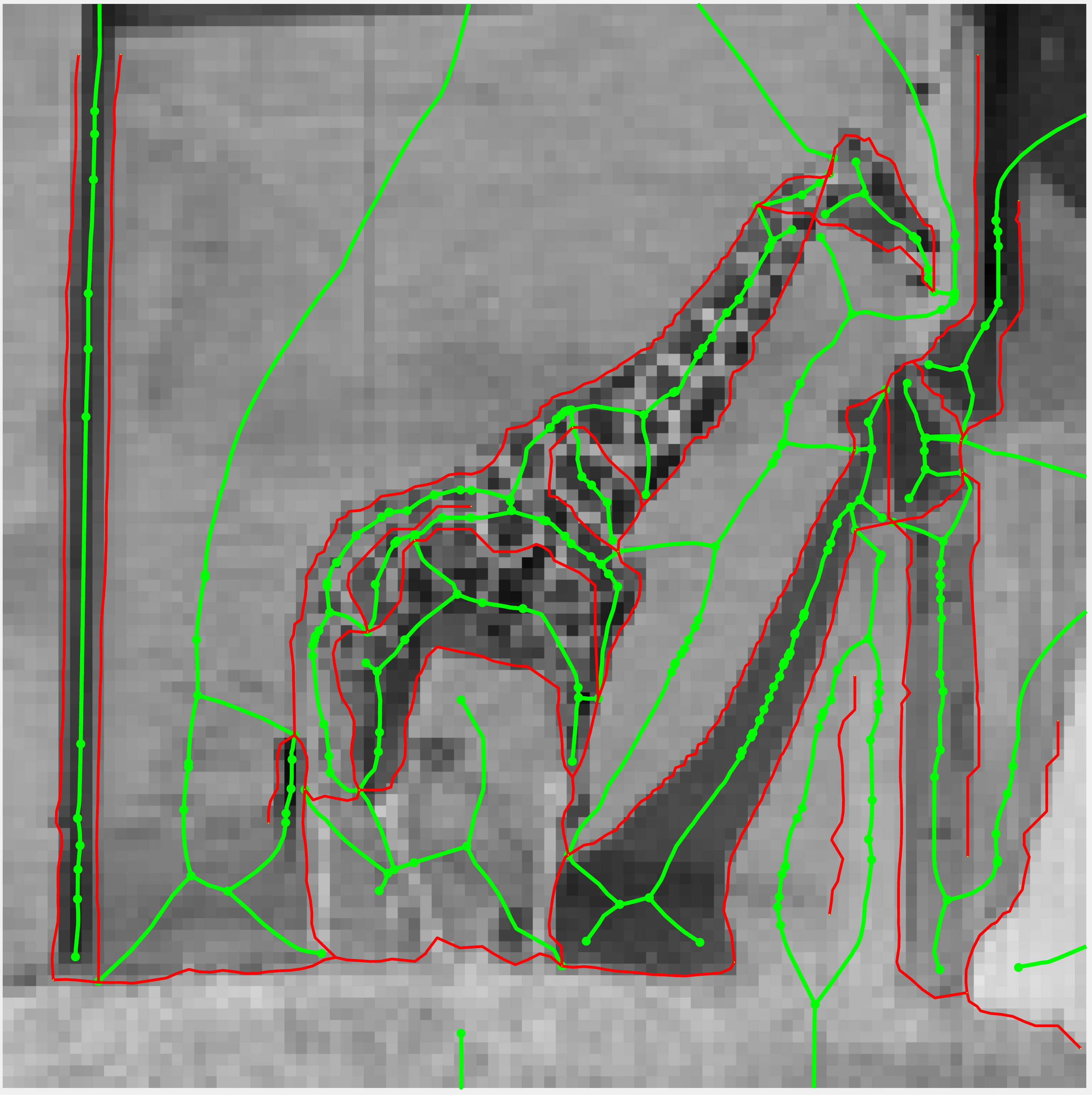} 
{\footnotesize\textit{\textcolor{white}{a)}}}\includegraphics[width=0.214\linewidth]{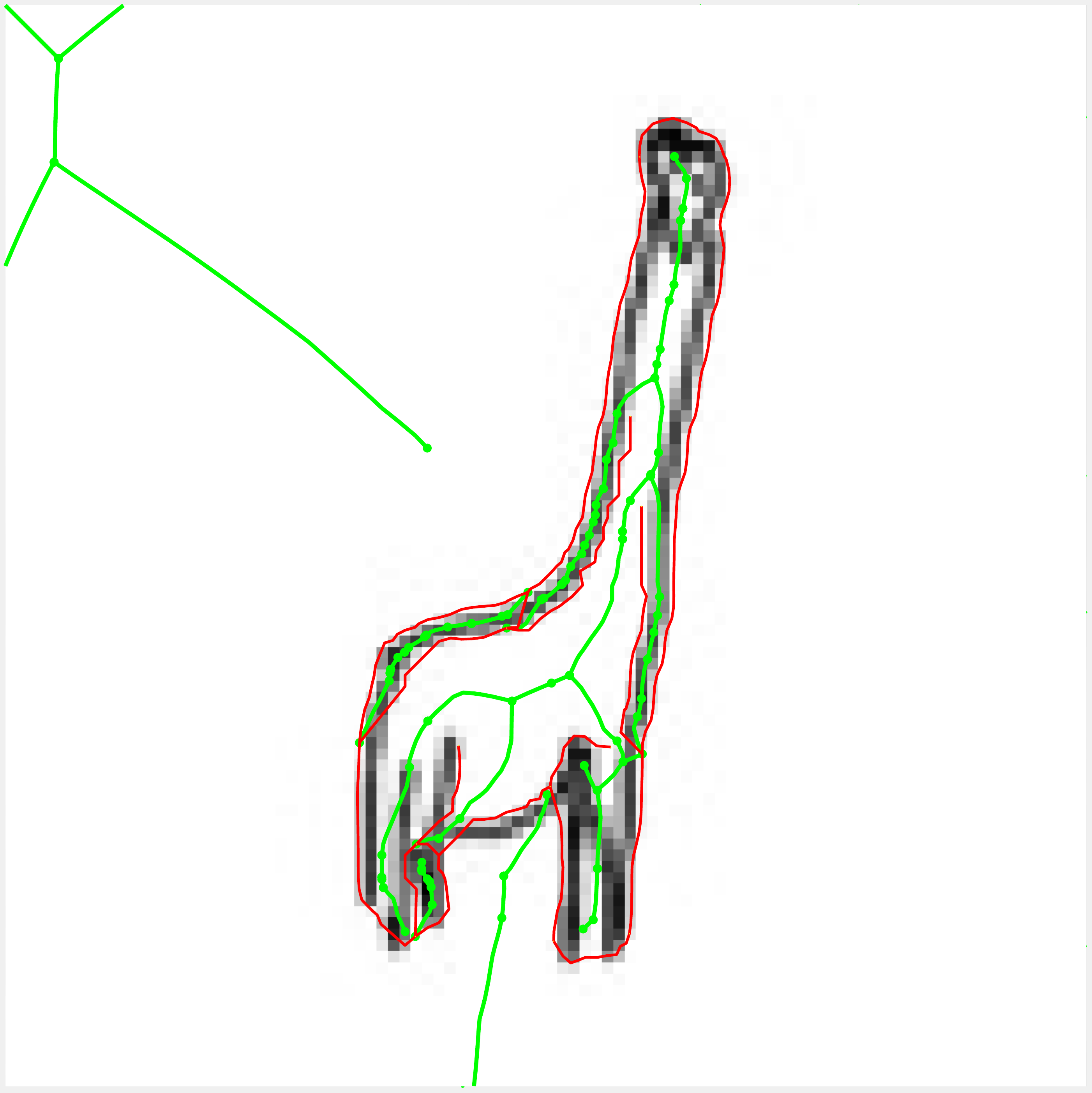}

\caption{a) Various images of a giraffe across the four domains from the PACS~\cite{Li:etal:ICCV17} dataset. b) The contour map on top of a grayscale version for clarity. c) The shock graph of the contour map colored in \textcolor{green}{green} (shock dynamics are not shown). We encourage readers to zoom in.}
\label{fig:pipeline}
\end{figure}


Our intuition for using the shock graph for domain generalization can be seen in in Figure~\ref{fig:pipeline}. Despite drastic differences in visual appearances we qualitatively observe that the shock graph~\ref{fig:pipeline}\textcolor{red}{c}, remains roughly the same even as we vary the domain. Furthermore, observe that the shock graphs of the various giraffes capture the common parts, distinctive neck and legs, shared among all giraffe exemplars.  We argue that this type of shape information is needed to bridge the domain gap. Also observe that texture cues such as the distinctive spots manifest as loops in the shock graph. Thus, the shock graph provides a rich descriptor of the global shape, and as shown in this giraffe case, the very distinctive surface texture. The shock graph is not limited to prominent shapes in an image. General scenes have a natural implicit shape or texture images contain repeating shape patterns which can also be captured by the shock graph.

Our quantitative framework then relies on defining an appropriate distance between shock graphs and subsequently using that in classification. Previous approaches defined a pairwise distance as the cost of transforming one shock graph to another using the edit distance algorithm~\cite{Sebastian:etal:Shocks:PAMI2004}.  We instead rely on the very popular state of the art approach of graph neural networks (GNN) to learn shock graph embeddings and subsequently classify them.


Graph neural networks are a very large and growing field, and a full discussion of the architectures, methods, and challenges are beyond the scope of the paper. We refer the reader to two excellent surveys~\cite{Wu:etal:Yu:Survey:GNN,Zhou:etal:Sun:Review:GNN} covering the field. We give a brief overview of GNNs and discuss the relevant aspects as it pertains to this paper. Identical to their Image CNN counterparts, GNNs mimic the same operations such as convolution, and pooling to extract information from the graph. For example, analogous to image based 2D convolution, graph convolution can be expressed as a weighted linear combination of neighborhood node information. We can make the weights (possibly bias) learnable coupled with non-linear activation functions to achieve a functionality similar to standard Image CNN convolutional layers. The stacking of such blocks, SoftMax of final layer output, and a cross-entropy loss function leads to a GNN architecture capable of performing node labeling, edge labeling, or graph classification. Differences in Convolution GNN architectures primarily alter how neighboring node information is aggregated (mean, max, sum, \etc) and how the neighborhood is defined (1 hop, 2 hops, \etc) for a node under consideration. Finally, in order to use shock graphs in a Convolutional GNN framework we need to augment each node or possibly edge with an initial fixed length feature vector. 

\begin{figure}[ht]
\center
{\footnotesize\textit{\textcolor{black}{a)}}}\includegraphics[width=0.22\textwidth]{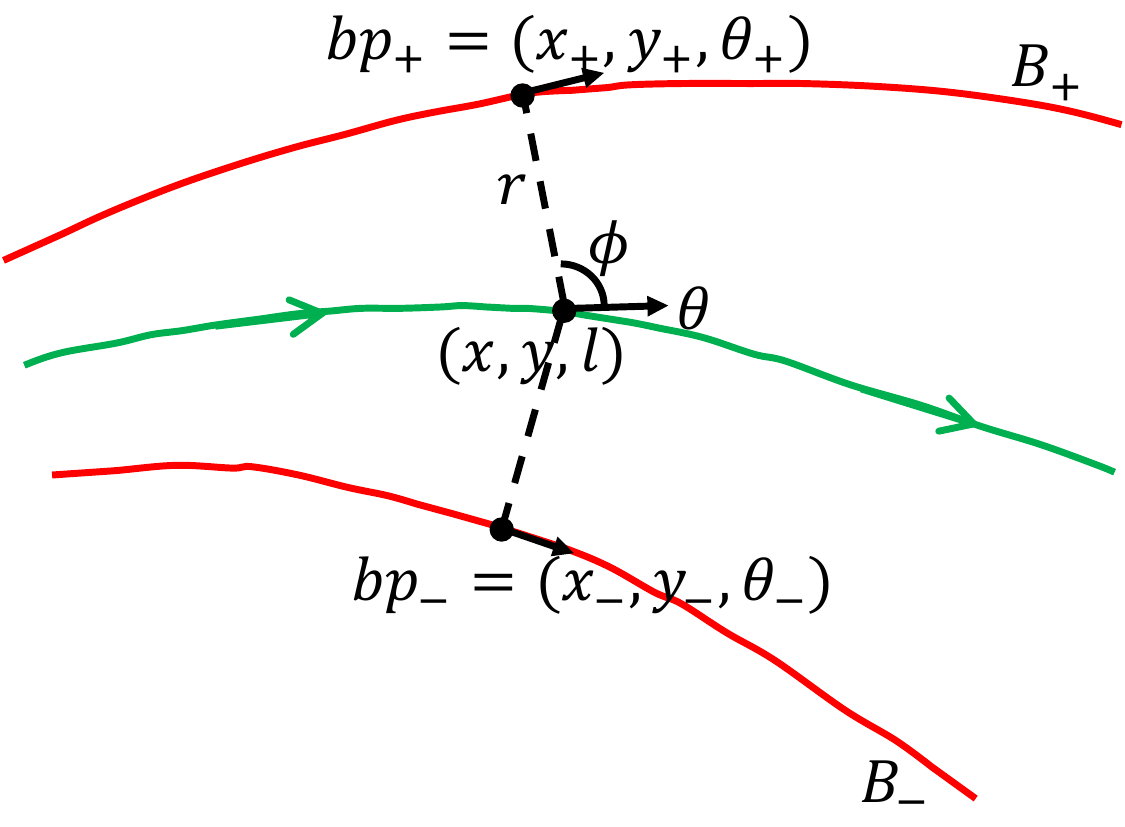}
{\footnotesize\textit{\textcolor{black}{b)}}}\includegraphics[width=0.22\textwidth]{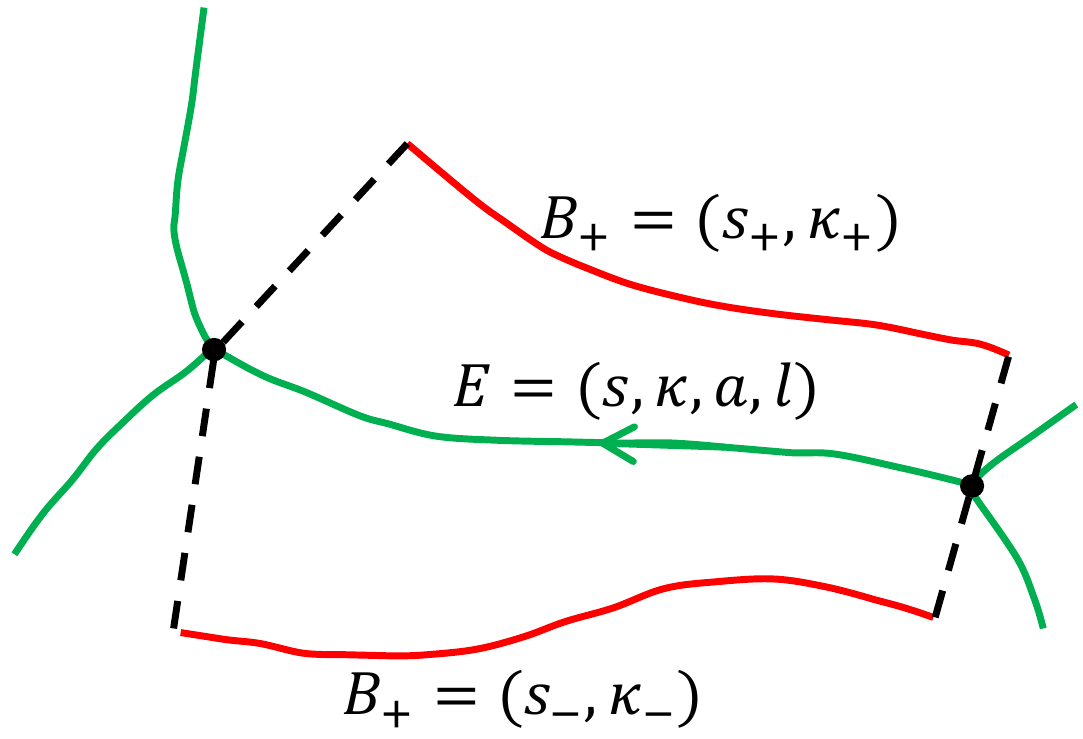}
\caption{a) First Order Geometric Attributes assigned to each Node. b) Edge Attributes composing geometric and differential geometric features. }
\label{fig:ne_attr}
\end{figure}

\noindent
\newline
 {\bf Feature selection:} Given a vertex $v$, graph convolutional layers generate the node's representation by aggregating its own features $h_v$, and its neighboring features $h_u$ where $u \in N(v)$ where $N$ represents the neighborhood. Stacking multiple graph convolution layers aggregates and extracts higher level node information. Initially, a starting set of features need to be assigned to each node. The analogy in Image CNN is the three channel RGB value attached to each pixel location of the input. We assigned each node and edge a set of first order geometric properties, Figure~\ref{fig:ne_attr}.

Observe in Figure~\ref{fig:ne_attr}\textcolor{red}{a} a typical degree two node. Each shock edge (bisector) is associated with the two contours, $B_+$ and $B_-$, that gave rise to it. Each node of an edge is also paired with the two interacting points on the curve, $bp_+$ and $bp_-$, which represent the boundary points tangent to the maximally inscribed circle of radius $r$. A feature vector, $h_u$, consisting of the following is assigned to each node:

\begin{equation}
  h_u=(x,y,r,l,\theta,\phi,\Psi)
\end{equation}

\noindent
where $(x,y)$ specifies the position of the shock node, $r$ is the shock radius, $l$ is a label of the type (sink,source,junction) of node, $\theta$ is the angle of the tangent to the shock curve, and $\phi$ represents the angle between the shock tangents and normals. $\Psi$ represents the set of oriented boundary points $bp_+$ and $bp_-$ which are endowed with a position $(x,y)$ and tangent $\theta$ respective to their associated boundaries.  For a degree two node then, we have six node features and six boundary point features leading to a feature vector of length twelve. A degree three node is associated with a triplet of interacting contours, leading to an increase in the number of shock edge tangents, shock edge normals, and boundary points to three.

Different from many of the graphs used in the GNN literature is that the shock graph is endowed with edge features. Figure~\ref{fig:ne_attr}\textcolor{red}{b} shows the local picture of each edge. Each shock edge is assigned a feature vector $h_e$ consisting of the following:

\begin{equation}
h_e=(s,\kappa,a,l,\Omega)
\end{equation}

\noindent
where $s$ refers to the arc-length, $\kappa$ the curvature of the shock edge, $a$ the area spanned by the shock edge and its two boundaries, $l$ is the label of the type (degenerate, semi-degenerate, regular)~\cite{Giblin:Kimia:IJCV03} of edge, and finally $\Omega$ refers to the properties associated with the pair of boundary curves leading rise to that particular shock edge. $\Omega$ represents the set of curve properties, arc-length $s$ and curvature $\kappa$, associated with each of the respective contours $B_+$ and $B_-$. This leads to an eight-dimensional vector composed of the four attributes attached to each shock edge coupled with the curve properties of the pair of boundaries. 

Our intuition for using edge features derives from the fact that the contours associated with each shock link provide a strong shape cue. For example, we would expect images of cars and chairs to exhibit many straights lines leading to low curvature values while natural images of dogs and giraffes to exhibit higher values due to the presence of wavy and circular smooth contours. How to properly exploit edge features is still an open problem in the GNN community, and while there is some minimal work on the problem, we found the simplest way of just attaching edge features to each node to be most effective. For example, a degree two node would have a node feature vector $h_u$ of length twelve coupled with two edge feature vectors $h_e$ of length eight leading to a twenty-eight dimensional feature vector. A difficulty with this is that nodes of varying degree will have a different feature vector length. To deal with this, we zero pad the feature vectors to a fixed length of fifty-eight so that it covers nodes with degrees varying from two to four.

\section{Methodology}

Our approach comprises two algorithmic steps: computation of the individual image shock graphs across a dataset and subsequently training/evaluating a graph neural network on this collection. Given an input image we initially compute an edge map using the Structured Forest approach of~\cite{Dollar:Zitnick:PAMI15}.  The orientation of the edges are corrected using the third-order approach of~\cite{Tamrakar:Kimia:ICCV07}. These edges are subsequently grouped into contour fragments (ordered set of edges) utilizing~\cite{Guo:etal:ECCV14}. The resulting set of contour fragments, typically on the order of eighty to hundred, are fed into the computational algorithm of~\cite{Tamrakar:Kimia:Shock} to produce the shock graph output.

The choice of Graph Neural Network architecture \ie type of graph convolution, number of layers, activation function, \etc is determined by experimentation on STL-10~\cite{Coates:etal:AISTATS11}. We picked this dataset as it is outside the domain generalization community and will not bias our results in any way. We utilized the Deep Graph Library~\cite{wang2019dgl} to evaluate three common graph architectures: Topological Adaptive Networks~\cite{Du:etal:CorR17} (TAG), Graph attention networks~\cite{Velickovic:etal:ICLR18} (GAT),  and graph convolutional networks~\cite{Kipf:Welling:ICLR17} (GCN)) and chose the architecture that performed best on our benchmark dataset. TAG achieved 58\%, while GAT  and GCN achieved 56.2\% and 54.3\%, respectively. Our final network depth is four layers, composed of three TAG convolutional layers and an ultimate linear layer coupled with SoftMax activation to produce class scores. The input layer transforms the initial 58-dimensional feature vector, described in Section~\ref{sec:sg}, for each node in the graph to a hidden dimension of 192 with subsequent TAG layers also having a hidden dimension of 192. Note that since we are doing graph classification, the node embeddings learned across multiple layers need to be merged into a single representation for the entire graph - we used simple mean pooling as an input into the final layer. Finally, following best practices in Image CNN architectures, we introduce batch norm and drop out between convolutional layers.

\section{Experiments}


We evaluated our framework on three popular image classification benchmarks: Colored MNIST~\cite{Arjovsky:etal:ARXIV20, Gulrajani:etal:ARXIV20}, PACS~\cite{Li:etal:ICCV17}, and VLCS~\cite{Fang:etal:ICCV13}. We follow the literature and perform ``leave-one-domain-out'' testing. In this paradigm we designate $N-1$ domains, where $N$ represents the number of domains, as ``source'' domains, and train our shock graph approach to completion on this collection. We then test this model on the ``leave-one'' ``target'' domain, with the expectation that a model with domain generalization capabilities will achieve similar performance on the source and the target datasets. We repeat this process across all source/target splits and report the image accuracy per domain and the average accuracy across all $N$ target domains. We compare our approach on these three datasets against a wide variety of state-of-the-art methods, Table~\ref{tbl:sota}. In each column of Table~\ref{tbl:sota} we have highlighted the best and second best methods in terms of classification performance in \textcolor{green}{green} and \textcolor{red}{red}, respectively.

There are two major challenges in fairly comparing our approach with the existing state-of-the-art methods: \emph{1)} the model complexity, and \emph{2)} the utility of pretrained ImageNet weights. For the first challenge, note that all existing methods are highly dependent on the backbone architecture being used - for example, the performance may greatly differ if a ResNet18 model is used compared to a ResNet50. While we are able to use a consistent backbone architecture for all of the Image CNN methods, our approach uses a GNN and thus cannot make use of the same architecture.  In particular, we note that most literature uses AlexNet (61M parameters), ResNet18 (11M parameters), or ResNet50 (25M parameters) backbones for their DG methods. Our method, in contrast, uses only 257k parameters, which has significant benefits not explicitly shown in the experimental metrics. These small neural networks coupled with the sparsity of the graphs allows us to achieve extremely fast inference times and process extremely large batch sizes on modern GPUs. So while we do incur some extra processing converting an image to graph, that is mostly offset by the major speed increase achieved during graph inference. Furthermore, when considering the results in Table~\ref{tbl:sota} if we were to enforce Image CNN architectures to have similar parameter complexity to our approach, its performance would be drastically poorer. 

The second major challenge in comparing our approach to the state-of-the-art domain generalization results is the use of pretrained ImageNet weights. It is very common for domain generalization methods to find powerful invariant deep features that are originally generated by training on the very large and diverse ImageNet dataset. While recent work such as~\cite{he2019rethinking} claim that CNN’s can achieve the same performance without ImageNet pretraining, the initial feature extraction from ImageNet provides a strong baseline for the invariant features that domain generalization methods discover, and in some cases are used directly for transfer learning. However, since our approach has not been trained on ImageNet, comparing it to domain generalization methods that utilize pretrained CNNs is not a fair comparison! This is because one method has benefited significantly from pretraining and the other has not. To mitigate this, we compare our results to the performance of state-of-the-art domain generalization methods with backbones starting from scratch (random weights). Training from scratch is NOT the best that these methods can do and the paper does not pretend it is, but it is necessary to have a fair comparison between the various approaches.



\begin{table*}[ht]
\begin{center}
    \begin{tabular}{ccccccc}
  \toprule

         &          &   \multicolumn{4}{c}{\bf Colored MNIST~\cite{Arjovsky:etal:ARXIV20, Gulrajani:etal:ARXIV20}} & \\
  \multicolumn{2}{c}{Method} & BackBone &  $\rho=0.1$ & $\rho=0.2$ & $\rho=0.9$ & Avg\\
  \hline

   \multicolumn{2}{c}{GroupDRO~\cite{Sagawa:etal:ICLR20}} & ResNet-18 (Scratch)        &     \textcolor{green}{72.7}   &       \textcolor{red}{73.1} &          10.2 &          52.0 \\
   \multicolumn{2}{c}{DeepALL~\cite{Li:etal:ICCV17}}    & ResNet-18 (Scratch)       &     71.7   &       73.0 &          10.3 &           \textcolor{red}{51.7} \\

   \multicolumn{2}{c}{MLDG~\cite{Li:mini:etal:AAAI18}}    & ResNet-18 (Scratch)        &     71.6   &       \textcolor{green}{73.2} &          10.1 &          51.6 \\
   \multicolumn{2}{c}{Self Challenge~\cite{Huang:etal:ECCV20}}  & ResNet-18 (Scratch)        &     66.6   &       65.7 &          \textcolor{red}{10.3} &           47.5 \\

  \hline
  \multicolumn{2}{c}{Our Approach}   & GNN-Tag &  \textcolor{red}{71.99} & 72.23 & \textcolor{green}{71.60} & \textcolor{green}{71.94} \\ 
    
  \hline
     
         &          & \multicolumn{4}{c}{\bf PACS Dataset~\cite{Li:etal:ICCV17}} & \\
  Method & BackBone &  Photo(P) & Art(A) & Cartoon(C) & Sketch(S) & Avg\\
  \hline

  Self Challenge~\cite{Huang:etal:ECCV20}& ResNet-18 (scratch)  & \textcolor{green}{55.02} & \textcolor{green}{42.38} & \textcolor{green}{53.28} & 37.15 & \textcolor{red}{46.95} \\
  GroupDRO~\cite{Sagawa:etal:ICLR20}      & ResNet-18 (scratch)  & 51.20 & 32.20 & 37.30 & 35.70 & 39.10 \\
  Episodic-DG~\cite{Li:mini:etal:ICCV19} & ResNet-18 (scratch)  & 41.13 & 29.83 & 42.15 & \textcolor{red}{37.69} & 37.70\\
  Jigsaw~\cite{Carlucci:mini:etal:CVPR19} & ResNet-18 (scratch)  & 42.34 & 30.37 & 45.65 & 29.14 & 36.66 \\
  MLDG~\cite{Li:mini:etal:AAAI18}         & ResNet-18 (scratch)  & 47.30 & 29.30 & 40.30 & 28.80 & 36.40 \\
  DeepALL~\cite{Li:etal:ICCV17}      & ResNet-50 (scratch)  & 50.80 & 29.10 & 39.20 & 24.00 & 35.80 \\

  DeepALL~\cite{Li:etal:ICCV17} & AlexNet (scratch)  & 21.53 & 22.14 & 18.11 &  8.940  & 17.68\\
  DeepALL~\cite{Li:etal:ICCV17} & ResNet-18 (scratch)  &  14.07 & 11.31 & 15.72 & 20.69 & 15.45 \\ 

  \hline
  Our Approach                       & GNN-Tag & \textcolor{red}{53.23} & \textcolor{red}{33.26} & \textcolor{red}{49.16} & \textcolor{green}{54.15} & \textcolor{green}{47.45}\\
  
  \hline
         &          & \multicolumn{4}{c}{\bf VLCS Dataset~\cite{Fang:etal:ICCV13}} & \\
  Method & BackBone &   Pascal(V) & LabelMe(L) & SUN(S) & Caltech(C) & Avg\\
  \hline
  MLDG~\cite{Li:mini:etal:AAAI18}         & ResNet-18 (scratch)  & 44.20 & 54.00 & 47.80 & \textcolor{red}{68.10}  & \textcolor{red}{53.50} \\
    Self Challenge~\cite{Huang:etal:ECCV20} & ResNet-18 (scratch) & 43.10 & \textcolor{green}{57.20}  & 45.20 &  66.00 & 52.90 \\
  GroupDRO~\cite{Sagawa:etal:ICLR20}      & ResNet-18 (scratch)  & \textcolor{red}{48.30} & 53.50 & \textcolor{red}{49.00} & 58.00 & 52.20 \\

  DeepALL~\cite{Li:etal:ICCV17} & ResNet-50 (scratch) & 45.30 & 49.20 & 43.60 & 64.40 & 50.60 \\
  DeepALL~\cite{Li:etal:ICCV17} & ResNet-18 (scratch)  & 45.35 & 46.42 & 39.85 & 63.67 & 48.82 \\
  DeepALL~\cite{Li:etal:ICCV17} & AlexNet (scratch)    & 43.25 & 47.61 & 38.51 & 62.51 & 47.97 \\

  \hline
  Our Approach                       & GNN-Tag & \textcolor{green}{49.15} & \textcolor{red}{55.49} & \textcolor{green}{52.13} &  \textcolor{green}{71.87} & \textcolor{green}{57.16} \\
  \bottomrule
\end{tabular}
\end{center}
\caption{Comparison of our approach (bottom row) to Domain Generalization methods. Each dataset column represents the target domain on which we test. The best classification accuracy is highlighted in \textcolor{green}{green} and the second best in \textcolor{red}{red}. The authors of Jigsaw~\cite{Carlucci:mini:etal:CVPR19} and  Episodic-DG~\cite{Li:mini:etal:ICCV19} only provided code to run on PACS.  Observe our approach is the best performing method across all datasets in average domain accuracy. }
\label{tbl:sota}
\end{table*}

\subsection{Colored MNIST}

\begin{figure}[!ht]

  \begin{center}
    \includegraphics[width=0.45\textwidth]{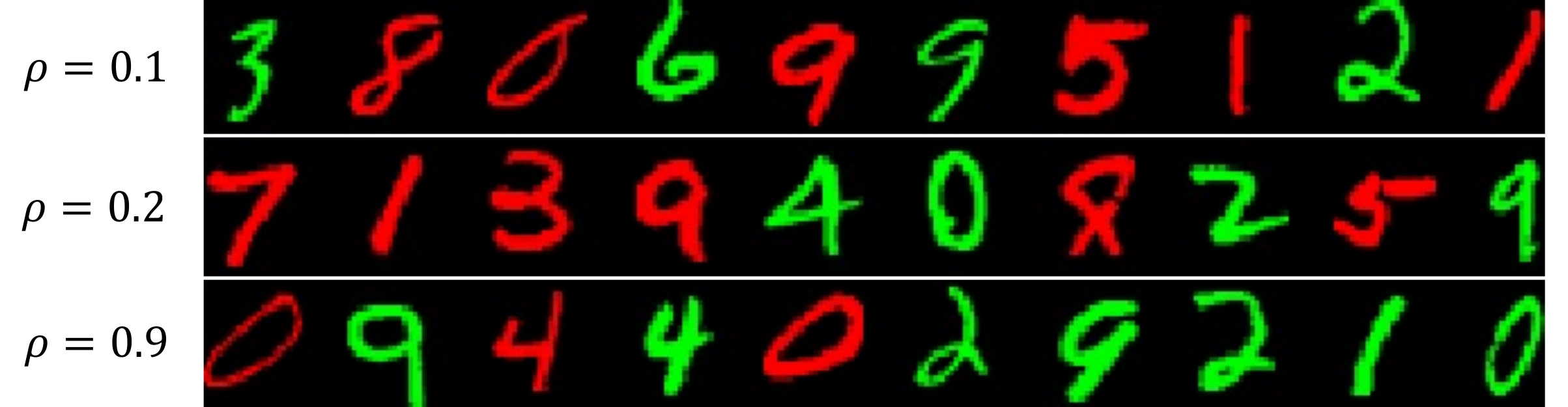}
    
\end{center}
\caption{a) Samples images from the Colored MNIST dataset. Note that the color of specific numbers significantly varies across domains. }
\label{fig:cmnist}
\end{figure}

We evaluated our approach on Colored MNIST~\cite{Arjovsky:etal:ARXIV20, Gulrajani:etal:ARXIV20} to empirically confirm that our approach utilizes the shape-based features when available, and also that existing methods fail to use shape-based features in favor of texture-based features. This dataset is a binary classification variant of the MNIST dataset and has three domains, Figure~\ref{fig:cmnist}, denoted $\rho = 0.1$, $\rho = 0.2$, and $\rho = 0.9$, each with varying degrees of how correlated the colors of the digits are with the true labels. For all domains, the shape of the digit has a $75\%$ correlation with the true label. While we leave the exact details of the construction of the dataset to the supplemental material, intuitively, the difference between these domains is that for $\rho = 0.1$, red is predictive of the label 0, while for $\rho = 0.9$, green is predictive of the label 0. Note that by construction, each individual domain has color more highly correlated with the label, which means that effective domain generalization requires that the shape is used (since the same color is not predictive of the same label across domains).


Observe that Image CNN based methods, Table~\ref{tbl:sota},  perform poorly on this dataset - in particular, they all perform reasonably
well when testing on the $\rho = 0.1$ and the $\rho = 0.2$ domain, but perform
significantly worse when testing on the $\rho = 0.9$ domain. This makes sense, as
training with a mixture of the $\rho = 0.9$ domain and either of the other two
domains necessitates the Image CNN based methods to focus on shape in order to achieve high performance  because color
is no longer highly correlated with the label. As expected, our graph
method focuses more on the shape-based cues and thus has consistent performance
across the domains. We also note that, for the Image CNN methods, the results were the same when pretrained ImageNet weights were used.

 
This experiment verifies that Image CNN based domain generalization methods do
neglect shape-based cues in favor of texture based cues when more predictive,
even when the shape-based cues may be more generalizable. Further, it validates
that our GNN domain generalization method works as intended and can focus on the
shape-based features which are more generalizable.

\subsection{PACS and VLCS} 

For a more traditional comparison, we use the heavily benchmarked VLCS and PACS datasets to evaluate our approach. VLCS~\cite{Fang:etal:ICCV13} includes images from four datasets (domains): PASCAL VOC2007 (V)~\cite{Everingham:etal:IJCV15}, LabelMe (L)~\cite{Russell:etal:IJCV08}, Caltech (C)~\cite{Fei-Fei:Fergus:Perona:CVPRW04} and SUN09 (S)~\cite{Choi:etal:CVPR10}. PACS~\cite{Li:etal:ICCV17} exhibits a larger cross domain shift than VLCS~\cite{Fang:etal:ICCV13}, and contains four domains covering Photo (P), Art Painting (A), Cartoon (C) and Sketch (S) images. Our results in Table~\ref{tbl:sota} show our method is consistently in the top two in performance across individual domains,  and our method is the top performing method when considering the mean classification performance across domains for both datasets. 

Observe that in PACS our performance on ``Art'' and ``Cartoon'' is worse than ``Sketch'' and ``Photo''. We attribute this to the fact that in the ``Sketch'' and ``Photos'' domains the role of shape and texture appear distinctly or not at all while in the ``Art'' and ``Cartoon'' domains texture is often itself represented as shape (internal contour patterns), thus confounding the role of shape. Further, observe that our performance in ``Sketch'' is vastly superior to all methods. This is a clear demonstration of the role of shape, as the black/white nature of sketches nullifies the role of appearance and methods generalizing from color-centric domains will suffer. We also notice in VLCS,  our performance on the SUN domain is the best performing method. This is somewhat surprising as SUN is a scene dataset, but shows that even in scenes the shock graph and GNN are able to recover an amorphous notion of shape that is predictive of the underlying class.

When assessing our results against other methods, we first look at how well we compare against DeepAll~\cite{Li:etal:ICCV17}, sometimes known as empirical risk minimization (ERM). DeepAll~\cite{Li:etal:ICCV17} trains an Image CNN with standard backbones (ResNet-18, ResNet-50, \emph{etc.}) on the source domains and evaluates its performance on the target domains. Since DeepAll is simple and is  one of the earliest approaches, it generally serves as a baseline to measure relative performance. Further, DeepALL is the closest analogue of our own work as we also do not do any complex training or evaluation procedures, but simply train our approach normally on the source domain and then evaluate on the target domain. Observe that our results exceed DeepALL (ResNet-50) by $32\%$ on PACS and $13\%$ on VLCS in terms of average domain accuracy regardless of backbone. Furthermore if we look at the individual target domains across PACS and VLCS we see our approach is better across the board. When comparing our approach against the rest of the methods, we notice that our approach is consistently the best or second best, where as the relative ranking of the other best performing methods differ according to the dataset (including Colored MNIST). This demonstrates that using the shape cue is more generalizable across different datasets than exclusively relying on learned appearance or texture cues. Finally, we note that our approach, despite not having the benefit of more advanced techniques like episodic training and meta-learning commonly employed by state-of-the-art approaches, is quite competitive across all domains and methods.

\section{Conclusion}


In this work, we merge classical computer vision techniques with
recent develops in graph neural networks in order to build a model
that is inherently robust to domain shift. We enforce the
inductive bias that our models focus on the shape of the images by
transforming an image to a graph by means of the shock graph, and then
train graph neural networks for subsequent classification tasks. With
no other changes, our proposed model shows competitive results in
domain generalization tasks. Contrary to existing methods, we bypass the difficulty that classical CNN's have in domain generalization by transforming the problem to the space of graphs, using invariant shape features to accomplish the task. An interesting area of future research would be to combine the strong feature extraction capabilities of CNN's with the shape features we obtain from the shock graph to achieve domain generalization. Finally, this work is scoped entirely to the task of image classification, but it could be extended to scene classification and object detection.

\section*{Acknowledgements}

The authors would like to thank the Tactical Intelligent Systems Group at JHU/APL for partially funding this research. Part of this research was conducted using computational resources and services at the Center for Computation and Visualization, Brown University. Benjamin Kimia gratefully acknowledges the support of NSF Award 1910530.

{\small
  \bibliographystyle{ieee_fullname}


\begin{thebibliography}{10}\itemsep=-1pt

\bibitem{Arjovsky:etal:ARXIV20}
Martin Arjovsky, Léon Bottou, Ishaan Gulrajani, and David Lopez-Paz.
\newblock Invariant risk minimization, 2020.

\bibitem{Azulay:Weiss:JMLR19}
Aharon Azulay and Yair Weiss.
\newblock Why do deep convolutional networks generalize so poorly to small
  image transformations?
\newblock {\em J. Mach. Learn. Res.}, 20:184:1--184:25, 2019.

\bibitem{Baker:etal:PLOS18}
Nicholas Baker, Hongjing Lu, Gennady Erlikhman, and Philip~J. Kellman.
\newblock Deep convolutional networks do not classify based on global object
  shape.
\newblock {\em PLOS Computational Biology}, 14:1--43, 12 2018.

\bibitem{Baker:etal:VR20}
Nicholas Baker, Hongjing Lu, Gennady Erlikhman, and Philip~J. Kellman.
\newblock Local features and global shape information in object classification
  by deep convolutional neural networks.
\newblock {\em Vision Research}, pages 46 -- 61, 2020.

\bibitem{Balaji:etal:NIPS2018}
Yogesh Balaji, Swami Sankaranarayanan, and Rama Chellappa.
\newblock Metareg: Towards domain generalization using meta-regularization.
\newblock In {\em Advances in Neural Information Processing Systems 31: Annual
  Conference on Neural Information Processing Systems 2018, NeurIPS 2018, 3-8
  December 2018, Montr{\'{e}}al, Canada}, pages 1006--1016, 2018.

\bibitem{Biederman:RBC}
Irving Biederman.
\newblock Recognition by components.
\newblock {\em Psych. Review}, 94:115--147, 1987.

\bibitem{Biederman:Ju:1988}
Irving Biederman and Ginny Ju.
\newblock Surface versus edge-based determinants of visual recognition.
\newblock {\em Cognitive Psychology}, 20:38--64, 1988.

\bibitem{Blum67Transformation}
H. Blum.
\newblock A transformation for extracting new descriptors of shape.
\newblock pages 362--380. M.I.T. Press, Cambridge, MA, U.S.A., 1967.
\newblock Proceedings of a Symposium held in Boston, MA, U.S.A.; November 1964.

\bibitem{Borji:ARXIV20}
Ali Borji.
\newblock Shape defense.
\newblock {\em CoRR}, abs/2008.13336, 2020.

\bibitem{Brendel:Bethge:ICLR19}
Wieland Brendel and Matthias Bethge.
\newblock Approximating cnns with bag-of-local-features models works
  surprisingly well on imagenet.
\newblock In {\em 7th International Conference on Learning Representations,
  {ICLR} 2019, New Orleans, LA, USA, May 6-9, 2019}. OpenReview.net, 2019.

\bibitem{Camaro:etal:CVPR20}
Charles{-}Olivier~Dufresne Camaro, Morteza Rezanejad, Stavros Tsogkas, Kaleem
  Siddiqi, and Sven~J. Dickinson.
\newblock Appearance shock grammar for fast medial axis extraction from real
  images.
\newblock In {\em Proceedings of the IEEE Computer Society Conference on
  Computer Vision and Pattern Recognition}, pages 14370--14379. IEEE, 2020.

\bibitem{carlini2017towards}
Nicholas Carlini and David Wagner.
\newblock Towards evaluating the robustness of neural networks.
\newblock In {\em 2017 ieee symposium on security and privacy (sp)}, pages
  39--57. IEEE, 2017.

\bibitem{Carlucci:etal:CVPR19}
Fabio~Maria Carlucci, Antonio D'Innocente, Silvia Bucci, Barbara Caputo, and
  Tatiana Tommasi.
\newblock Domain generalization by solving jigsaw puzzles.
\newblock In {\em Proceedings of the IEEE Computer Society Conference on
  Computer Vision and Pattern Recognition}, pages 2229--2238. IEEE, 2019.

\bibitem{Carlucci:mini:etal:CVPR19}
Fabio~Maria Carlucci, Antonio D'Innocente, Silvia Bucci, Barbara Caputo, and
  Tatiana Tommasi.
\newblock Domain generalization by solving jigsaw puzzles.
\newblock In {\em CVPR}, 2019.

\bibitem{Choi:etal:CVPR10}
Myung~Jin Choi, Joseph~J. Lim, Antonio Torralba, and Alan~S. Willsky.
\newblock Exploiting hierarchical context on a large database of object
  categories.
\newblock In {\em Proceedings of the IEEE Computer Society Conference on
  Computer Vision and Pattern Recognition}, pages 129--136, San Francisco,
  California, USA, 2010. IEEE Computer Society Press.

\bibitem{Coates:etal:AISTATS11}
Adam Coates, Andrew~Y. Ng, and Honglak Lee.
\newblock An analysis of single-layer networks in unsupervised feature
  learning.
\newblock In {\em AISTATS2011}, 2011.

\bibitem{Dollar:Zitnick:PAMI15}
Piotr Doll{\'{a}}r and C.~Lawrence Zitnick.
\newblock Fast edge detection using structured forests.
\newblock {\em {IEEE} Trans. Pattern Anal. Mach. Intell.}, 37(8):1558--1570,
  2015.

\bibitem{Dou:etal:NIPS19}
Qi Dou, Daniel~Coelho de Castro, Konstantinos Kamnitsas, and Ben Glocker.
\newblock Domain generalization via model-agnostic learning of semantic
  features.
\newblock In {\em Advances in Neural Information Processing Systems 32: Annual
  Conference on Neural Information Processing Systems 2019, NeurIPS 2019, 8-14
  December 2019, Vancouver, BC, Canada}, pages 6447--6458, 2019.

\bibitem{Du:etal:CorR17}
Jian Du, Shanghang Zhang, Guanhang Wu, Jos{\'{e}} M.~F. Moura, and Soummya Kar.
\newblock Topology adaptive graph convolutional networks.
\newblock {\em CoRR}, 2017.

\bibitem{Everingham:etal:IJCV15}
Mark Everingham, S.~M.~Ali Eslami, Luc~Van Gool, Christopher K.~I. Williams,
  John~M. Winn, and Andrew Zisserman.
\newblock The pascal visual object classes challenge: {A} retrospective.
\newblock {\em International Journal of Computer Vision}, 111(1):98--136, 2015.

\bibitem{Fang:etal:ICCV13}
Chen Fang, Ye Xu, and Daniel~N. Rockmore.
\newblock Unbiased metric learning: On the utilization of multiple datasets and
  web images for softening bias.
\newblock In {\em Proceedings of the IEEE International Conference on Computer
  Vision}, pages 1657--1664. IEEE, 2013.

\bibitem{Fei-Fei:Fergus:Perona:CVPRW04}
Li Fei-Fei, R. Fergus, and Pietro Perona.
\newblock Learning generative visual models from few training examples: an
  incremental {B}ayesian approach tested on 101 object categories.
\newblock In {\em Proc. of IEEE CVPR Workshop on Generative Model Based
  Vision.}, 2004.

\bibitem{Felzenszwalb:Schwartz:CVPR:2007}
Pedro~F. Felzenszwalb and Joshua~D. Schwartz.
\newblock Hierarchical matching of deformable shapes.
\newblock In {\em Proceedings of the IEEE Computer Society Conference on
  Computer Vision and Pattern Recognition}. IEEE Computer Society, 2007.

\bibitem{Fin:etal:ICML17}
Chelsea Finn, Pieter Abbeel, and Sergey Levine.
\newblock Model-agnostic meta-learning for fast adaptation of deep networks.
\newblock In {\em Proceedings of the 34th International Conference on Machine
  Learning}, pages 1126--1135, 2017.

\bibitem{Geirhos:etal:ICLR19}
Robert Geirhos, Patricia Rubisch, Claudio Michaelis, Matthias Bethge, Felix~A.
  Wichmann, and Wieland Brendel.
\newblock Imagenet-trained cnns are biased towards texture; increasing shape
  bias improves accuracy and robustness.
\newblock In {\em 7th International Conference on Learning Representations,
  {ICLR} 2019, New Orleans, LA, USA, May 6-9, 2019}. OpenReview.net, 2019.

\bibitem{Ghifary:etal:CVPR15}
Muhammad Ghifary, W~Bastiaan Kleijn, Mengjie Zhang, and David Balduzzi.
\newblock Domain generalization for object recognition with multi-task
  autoencoders.
\newblock In {\em Proceedings of the IEEE Computer Society Conference on
  Computer Vision and Pattern Recognition}, page 2551–2559. IEEE, 2015.

\bibitem{Giblin:Kimia:Reconstruction:PAMI03}
Peter~J. Giblin and Benjamin~B. Kimia.
\newblock On the intrinsic reconstruction of shape from its symmetries.
\newblock {\em IEEE Trans. Pattern Anal. Mach. Intell.}, 25(7):895--911, July
  2003.

\bibitem{Giblin:Kimia:IJCV03}
Peter.~J. Giblin and Benjamin.~B. Kimia.
\newblock On the local form and transitions of symmetry sets, medial axes, and
  shocks.
\newblock {\em International Journal of Computer Vision}, 54(Issue
  1-3):143--157, August 2003.

\bibitem{Giblin:Kimia:MedialBook07}
Peter~J. Giblin and Benjamin~B. Kimia.
\newblock Local forms and transitions of the medial axis.
\newblock In Kaleem Siddiqi and Stephen Pizer, editors, {\em Medial
  Representations: Mathematics, Algorithms and Applications}, pages 37--68.
  Kluwer Academic Publishers, 2008.

\bibitem{goodfellow2014explaining}
Ian~J Goodfellow, Jonathon Shlens, and Christian Szegedy.
\newblock Explaining and harnessing adversarial examples.
\newblock {\em arXiv preprint arXiv:1412.6572}, 2014.

\bibitem{Gulrajani:etal:ARXIV20}
Ishaan Gulrajani and David Lopez-Paz.
\newblock In search of lost domain generalization, 2020.

\bibitem{Guo:etal:ECCV14}
Yuliang Guo, Naman Kumar, Maruthi Narayanan, and Benjamin~B. Kimia.
\newblock A multi-stage approach to curve extraction.
\newblock In {\em Proceedings of European Conference on Computer Vision},
  Lecture Notes in Computer Science, pages 663--678. Springer, 2014.

\bibitem{he2019rethinking}
Kaiming He, Ross Girshick, and Piotr Doll{\'a}r.
\newblock Rethinking imagenet pre-training.
\newblock In {\em Proceedings of the IEEE International Conference on Computer
  Vision}, pages 4918--4927, 2019.

\bibitem{Hendrycks:etal:ICLR19}
Dan Hendrycks and Thomas Dietterich.
\newblock Benchmarking neural network robustness to common corruptions and
  perturbations.
\newblock {\em Proceedings of the International Conference on Learning
  Representations}, 2019.

\bibitem{Hosseini:etal:CVPRW18}
Hossein Hosseini, Baicen Xiao, Mayoore Jaiswal, and Radha Poovendran.
\newblock Assessing shape bias property of convolutional neural networks.
\newblock In {\em 2018 {IEEE} Conference on Computer Vision and Pattern
  Recognition Workshops, {CVPR} Workshops 2018, Salt Lake City, UT, USA, June
  18-22, 2018}, pages 1923--1931, 2018.

\bibitem{Huang:etal:ECCV20}
Zeyi Huang, Haohan Wang, Eric~P. Xing, and Dong Huang.
\newblock Self-challenging improves cross-domain generalization.
\newblock In {\em ECCV}, 2020.

\bibitem{Khosla:etal:ECCV12}
Aditya Khosla, Tinghui Zhou, Tomasz Malisiewicz, Alexei~A. Efros, and Antonio
  Torralba.
\newblock Undoing the damage of dataset bias.
\newblock In {\em Proceedings of European Conference on Computer Vision},
  Lecture Notes in Computer Science, pages 158--171. Springer, 2012.

\bibitem{Kimia:etal:ECCV:Book}
Benjamin~B. Kimia, Allen~R. Tannenbaum, and Steven~W. Zucker.
\newblock Toward a computational theory of shape: An overview.
\newblock In Olivier~D. Faugeras, editor, {\em Proceedings of European
  Conference on Computer Vision}, volume 427 of {\em Lecture Notes in Computer
  Science}, pages 402--407. Springer, 1990.

\bibitem{Kimia:etal:Shape:Series:I}
Benjamin~B. Kimia, Allen~R. Tannenbaum, and Steven~W. Zucker.
\newblock Shapes, shocks, and deformations,~{I}: The components of shape and
  the reaction-diffusion space.
\newblock {\em International Journal of Computer Vision}, 15(3):189--224, 1995.

\bibitem{Kipf:Welling:ICLR17}
Thomas~N. Kipf and Max Welling.
\newblock Semi-supervised classification with graph convolutional networks.
\newblock In {\em ICLR2017}, 2017.

\bibitem{Kreiegeskorte:etal:ARVS15}
Nikolaus Kriegeskorte.
\newblock Deep neural networks: A new framework for modeling biological vision
  and brain information processing.
\newblock {\em Annual Review of Vision Science}, 1(1):417--446, 2015.

\bibitem{Krizhevsky09learningmultiple}
Alex Krizhevsky.
\newblock Learning multiple layers of features from tiny images.
\newblock Technical report, 2009.

\bibitem{LeCun:etal:NATURE15}
Yann LeCun, Yoshua Bengio, and Geoffrey~E. Hinton.
\newblock Deep learning.
\newblock {\em Nature}, 521(7553):436--444, 2015.

\bibitem{Li:etal:ICCV17}
Da Li, Yongxin Yang, Yi{-}Zhe Song, and Timothy~M. Hospedales.
\newblock Deeper, broader and artier domain generalization.
\newblock In {\em Proceedings of the IEEE International Conference on Computer
  Vision}, pages 5543--5551. IEEE, 2017.

\bibitem{Li:etal:AAAI18}
Da Li, Yongxin Yang, Yi{-}Zhe Song, and Timothy~M. Hospedales.
\newblock Learning to generalize: Meta-learning for domain generalization.
\newblock In {\em Proceedings of the Thirty-Second {AAAI} Conference on
  Artificial Intelligence, New Orleans, Louisiana, USA, February 2-7, 2018},
  pages 3490--3497. AAAI Press, 2018.

\bibitem{Li:mini:etal:AAAI18}
Da Li, Yongxin Yang, Yi{-}Zhe Song, and Timothy~M. Hospedales.
\newblock Learning to generalize: Meta-learning for domain generalization.
\newblock In {\em AAAI}, 2018.

\bibitem{Li:etal:ICCV19}
Da Li, Jianshu Zhang, Yongxin Yang, Cong Liu, Yi{-}Zhe Song, and Timothy~M.
  Hospedales.
\newblock Episodic training for domain generalization.
\newblock In {\em Proceedings of the IEEE International Conference on Computer
  Vision}, pages 1446--1455. IEEE, 2019.

\bibitem{Li:mini:etal:ICCV19}
Da Li, Jianshu Zhang, Yongxin Yang, Cong Liu, Yi{-}Zhe Song, and Timothy~M.
  Hospedales.
\newblock Episodic training for domain generalization.
\newblock In {\em ICCV}, 2019.

\bibitem{Li:etal:CVPR18}
Haoliang Li, Sinno~Jialin Pan, Shiqi Wang, and Alex~C. Kot.
\newblock Domain generalization with adversarial feature learning.
\newblock In {\em Proceedings of the IEEE Computer Society Conference on
  Computer Vision and Pattern Recognition}, pages 5400--5409. IEEE, 2018.

\bibitem{Li:etal:ECCV18}
Ya Li, Xinmei Tian, Mingming Gong, Yajing Liu, Tongliang Liu, Kun Zhang, and
  Dacheng Tao.
\newblock Deep domain generalization via conditional invariant adversarial
  networks.
\newblock In {\em Proceedings of European Conference on Computer Vision},
  Lecture Notes in Computer Science, pages 647--663. Springer, 2016.

\bibitem{Li:etal:ICLR21}
Yingwei Li, Qihang Yu, Mingxing Tan, Jieru Mei, Peng Tang, Wei Shen, Alan~L.
  Yuille, and Cihang Xie.
\newblock Shape-texture debiased neural network training.
\newblock In {\em 9th International Conference on Learning Representations,
  {ICLR} 2021, Virtual Event, Austria, May 3-7, 2021}, 2021.

\bibitem{Malhotra:Bowers:CS19}
Gaurav Malhotra and Jeff Bowers.
\newblock The contrasting roles of shape in human vision and convolutional
  neural networks.
\newblock In {\em Proceedings of the 41th Annual Meeting of the Cognitive
  Science Society, CogSci 2019: Creativity + Cognition + Computation, Montreal,
  Canada, July 24-27, 2019}, pages 2261--2267, 2019.

\bibitem{Marr:Visual:1978}
David Marr.
\newblock {\em Representation Visual Information}.
\newblock 1978.

\bibitem{moosavi2016deepfool}
Seyed-Mohsen Moosavi-Dezfooli, Alhussein Fawzi, and Pascal Frossard.
\newblock Deepfool: a simple and accurate method to fool deep neural networks.
\newblock In {\em Proceedings of the IEEE conference on computer vision and
  pattern recognition}, pages 2574--2582, 2016.

\bibitem{Mori:etal:PAMI05}
Greg Mori, Serge~J. Belongie, and Jitendra Malik.
\newblock Efficient shape matching using shape contexts.
\newblock {\em IEEE Trans. Pattern Anal. Mach. Intell.}, 27(11):1832--1837,
  2005.

\bibitem{Motiian:etal:ICCV17}
Saeid Motiian, Marco Piccirilli, Donald~A. Adjeroh, and Gianfranco Doretto.
\newblock Unified deep supervised domain adaptation and generalization.
\newblock In {\em Proceedings of the IEEE International Conference on Computer
  Vision}, pages 5716--5726. IEEE, 2017.

\bibitem{Munandet:etal:ICML13}
Krikamol Muandet, David Balduzzi, and Bernhard Sch{\"{o}}lkopf.
\newblock Domain generalization via invariant feature representation.
\newblock In {\em Proceedings of the 30th International Conference on Machine
  Learning, {ICML} 2013, Atlanta, GA, USA, 16-21 June 2013}, pages 10--18.
  JMLR.org, 2013.

\bibitem{Narayanan:Kimia:ECCV12}
Maruthi Narayanan and Benjamin~B. Kimia.
\newblock Bottom-up perceptual organization of images into object part
  hypotheses.
\newblock In {\em Proceedings of European Conference on Computer Vision},
  Lecture Notes in Computer Science, pages 257--271. Springer, 2012.

\bibitem{Ozcanli:Kimia:BMVC07}
Ozge~C. Ozcanli and Benjamin~B. Kimia.
\newblock Generic object recognition via shock patch fragments.
\newblock In Nasir~M. Rajpoot and Abhir Bhalerao, editors, {\em Proceedings of
  the British Machine Vision Conference}, pages 1030--1039, Coventry, USA,
  September 10-13 2007. Warwick Print.

\bibitem{Qiao:etal:CVPR20}
Fengchun Qiao, Long Zhao, and Xi Peng.
\newblock Learning to learn single domain generalization.
\newblock In {\em Proceedings of the IEEE Computer Society Conference on
  Computer Vision and Pattern Recognition}, pages 12553--12562. IEEE, 2020.

\bibitem{Ritter:etal:ICML17}
Samuel Ritter, David G.~T. Barrett, Adam Santoro, and Matt~M. Botvinick.
\newblock Cognitive psychology for deep neural networks: {A} shape bias case
  study.
\newblock In {\em Proceedings of the 34th International Conference on Machine
  Learning, {ICML} 2017, Sydney, NSW, Australia, 6-11 August 2017}, volume~70
  of {\em Proceedings of Machine Learning Research}, pages 2940--2949.

\bibitem{Russell:etal:IJCV08}
Bryan~C. Russell, Antonio Torralba, Kevin~P. Murphy, and William~T. Freeman.
\newblock Labelme: {A} database and web-based tool for image annotation.
\newblock {\em Int. J. Comput. Vis.}, 77(1-3):157--173, 2008.

\bibitem{Sagawa:etal:ICLR20}
Shiori Sagawa, Pang~Wei Koh, Tatsunori~B. Hashimoto, and Percy Liang.
\newblock Distributionally robust neural networks.
\newblock In {\em ICLR}, 2020.

\bibitem{Sebastian:etal:Shocks:PAMI2004}
Thomas Sebastian, Philip Klein, and Benjamin Kimia.
\newblock Recognition of shapes by editing their shock graphs.
\newblock {\em IEEE Trans. Pattern Anal. Mach. Intell.}, 26:551--571, May 2004.

\bibitem{Shankar:etal:ICLR18}
Shiv Shankar, Vihari Piratla, Soumen Chakrabarti, Siddhartha Chaudhuri, Preethi
  Jyothi, and Sunita Sarawagi.
\newblock Generalizing across domains via cross-gradient training.
\newblock In {\em 6th International Conference on Learning Representations,
  {ICLR} 2018, Vancouver, BC, Canada, April 30 - May 3, 2018}. OpenReview.net,
  2018.

\bibitem{Shi:etal:ICML20}
Baifeng Shi, Dinghuai Zhang, Qi Dai, Zhanxing Zhu, Yadong Mu, and Jingdong
  Wang.
\newblock Informative dropout for robust representation learning: {A}
  shape-bias perspective.
\newblock In {\em Proceedings of the 37th International Conference on Machine
  Learning, {ICML} 2020, 13-18 July 2020, Virtual Event}, pages 8828--8839.
  {PMLR}, 2020.

\bibitem{szegedy2013intriguing}
Christian Szegedy, Wojciech Zaremba, Ilya Sutskever, Joan Bruna, Dumitru Erhan,
  Ian Goodfellow, and Rob Fergus.
\newblock Intriguing properties of neural networks.
\newblock {\em arXiv preprint arXiv:1312.6199}, 2013.

\bibitem{Tamrakar:Kimia:ICCV07}
Amir Tamrakar and Benjamin~B. Kimia.
\newblock No grouping left behind: From edges to curve fragments.
\newblock In {\em Proceedings of the IEEE International Conference on Computer
  Vision}, Rio de Janeiro, Brazil, October 2007. IEEE Computer Society.

\bibitem{Tamrakar:Kimia:Shock}
Amir Tamrakar and Benjamin~B. Kimia.
\newblock Intrinsic shock representation and shock computation from points,
  lines and circular arcs in an eulerian-lagrangian framework.
\newblock Technical Report LEMS-199, Laboratory for Engineering Man/Machine
  Systems, Brown University, 2008.

\bibitem{Trinh:Kimia:IJCV11}
Nhon~H. Trinh and Benjamin~B. Kimia.
\newblock {\it Skeleton Search}: Category-specific object recognition and
  segmentation using a skeletal shape model.
\newblock {\em International Journal of Computer Vision}, 94(2):215--240, 2011.

\bibitem{Tsogkas:Dickinson:ICCV17}
Stavros Tsogkas and Sven~J. Dickinson.
\newblock {AMAT:} medial axis transform for natural images.
\newblock In {\em Proceedings of the IEEE International Conference on Computer
  Vision}, pages 2727--2736. IEEE, 2017.

\bibitem{Velickovic:etal:ICLR18}
Petar Velickovic, Guillem Cucurull, Arantxa Casanova, Adriana Romero, Pietro
  Li{\`{o}}, and Yoshua Bengio.
\newblock Graph attention networks.
\newblock In {\em ICLR2018}, 2018.

\bibitem{Vidal:etal:MedialAxis:ICPR00}
S.F. Vidal, E. Bardinet, G. Malandain, S. Damas, and N.P. de~la Blanca~Capilla.
\newblock Object representation and comparison inferred from its medial axis.
\newblock In {\em ICPR00}, pages Vol I: 712--715, 2000.

\bibitem{Volpi:etal:NIPS18}
Riccardo Volpi, Hongseok Namkoong, Ozan Sener, John~C. Duchi, Vittorio Murino,
  and Silvio Savarese.
\newblock Generalizing to unseen domains via adversarial data augmentation.
\newblock In {\em Advances in Neural Information Processing Systems 31: Annual
  Conference on Neural Information Processing Systems 2018, NeurIPS 2018, 3-8
  December 2018, Montr{\'{e}}al, Canada}, pages 5339--5349, 2018.

\bibitem{Wang:etal:NIPS19}
Haohan Wang, Songwei Ge, Zachary~C. Lipton, and Eric~P. Xing.
\newblock Learning robust global representations by penalizing local predictive
  power.
\newblock pages 10506--10518, NIPS2019.

\bibitem{wang2019dgl}
Minjie Wang, Da Zheng, Zihao Ye, Quan Gan, Mufei Li, Xiang Song, Jinjing Zhou,
  Chao Ma, Lingfan Yu, Yu Gai, Tianjun Xiao, Tong He, George Karypis, Jinyang
  Li, and Zheng Zhang.
\newblock Deep graph library: A graph-centric, highly-performant package for
  graph neural networks.
\newblock {\em arXiv preprint arXiv:1909.01315}, 2019.

\bibitem{Wu:etal:Yu:Survey:GNN}
Zonghan Wu, Shirui Pan, Fengwen Chen, Guodong Long, Chengqi Zhang, and
  Philip~S. Yu.
\newblock A comprehensive survey on graph neural networks.
\newblock {\em CoRR}, abs/1901.00596, 2019.

\bibitem{Zeiler:Fergus:ECCV14}
Matthew~D. Zeiler and Rob Fergus.
\newblock Visualizing and understanding convolutional networks.
\newblock In {\em Proceedings of European Conference on Computer Vision},
  Lecture Notes in Computer Science, pages 818--833. Springer, 2014.

\bibitem{Zhang:etal:CVPR20}
Yuwei Zhang, Peng Zhang, Chun Yuan, and Zhi Wang.
\newblock Texture and shape biased two-stream networks for clothing
  classification and attribute recognition.
\newblock In {\em Proceedings of the IEEE Computer Society Conference on
  Computer Vision and Pattern Recognition}, pages 13535--13544. IEEE, 2020.

\bibitem{Zhou:etal:Sun:Review:GNN}
Jie Zhou, Ganqu Cui, Zhengyan Zhang, Cheng Yang, Zhiyuan Liu, and Maosong Sun.
\newblock Graph neural networks: {A} review of methods and applications.
\newblock {\em CoRR}, abs/1812.08434, 2018.

\end{thebibliography}

\end{document}